\documentclass[final,3p,times,twocolumn]{elsarticle}

\usepackage{amssymb}
\usepackage{amsmath}
\usepackage{multirow}
\usepackage{hyperref}
\usepackage{algorithm}
\usepackage{algpseudocode}
\usepackage{graphicx}
\usepackage{makecell}
\usepackage{bbding}
\usepackage{colortbl}

\newcommand{\eqautoref}[1]{\hyperref[#1]{(\ref*{#1})}}

\biboptions{sort&compress}

\journal{Pattern Recognition}

\begin{document}

\begin{frontmatter}

\title{PID: Physics-Informed Diffusion Model for Infrared Image Generation}

\author{%
  Fangyuan Mao\(^{1,2}\), Jilin Mei\(^{1}\), Shun Lu\(^{1,2}\), Fuyang Liu\(^{1,2}\), Liang Chen\(^{1}\), Fangzhou Zhao\(^{1}\), Yu Hu\(^{1,2}\) \\
  \(^{1}\) Research Center for Intelligent Computing Systems,\\
  Institute of Computing Technology, Chinese Academy of Sciences\\
  \(^{2}\) School of Computer Science and Technology, University of Chinese Academy of Sciences\\
  \texttt{\{maofangyuan23s,meijilin,lushun19s,chenliang,zhaofangzhou,huyu\}@ict.ac.cn} \\
}

\begin{abstract}

Infrared imaging technology has gained notable attention for its reliable sensing ability in low visibility conditions, prompting numerous studies to convert the abundant RGB images to infrared images. 
However, most existing image translation methods treat infrared images as a stylistic variation, neglecting the fundamental principles of physics, which limits their practical application. 
To address these issues, we propose a \textbf{P}hysics-\textbf{I}nformed \textbf{D}iffusion (PID) model for translating RGB images to infrared images that adhere to physical laws. 
Our method leverages the iterative optimization of the diffusion model and incorporates strong physical constraints based on prior knowledge of infrared laws during training. This approach enhances the similarity between translated infrared images and the real infrared domain without increasing extra inference parameters. Experimental results demonstrate that PID significantly outperforms existing state-of-the-art methods. Our code is available at \href{https://github.com/fangyuanmao/PID}{https://github.com/fangyuanmao/PID}.
\end{abstract}


\end{frontmatter}

\section{Introduction}
\label{Introduction}

Infrared images offer strong anti-interference capabilities and demonstrate resilience in complex weather conditions like night, rain, dust, and snow. In fields like robotics and autonomous driving, there is a growing demand of infrared images for perception tasks~\cite{kou2023infrared}. 
However, acquiring infrared images requires specialized devices, which makes the availability of open-source infrared datasets lag behind that of RGB image datasets. To address this data shortage, self-supervised training~\cite{Zhang2023UnleashingTP} and generative networks~\cite{Alzubaidi2023ASO} have become common approaches. Previous studies~\cite{Kniaz2018ThermalGANMC,zkanoglu2022InfraGANAG,Lee2023EdgeguidedMR} have tried to translate RGB images to infrared images to augment infrared datasets. Despite remarkable advancements in recent visible-to-infrared researches, two major limitations remain.

\begin{figure*}[!ht]
        \centering
        \includegraphics[width=1\linewidth]{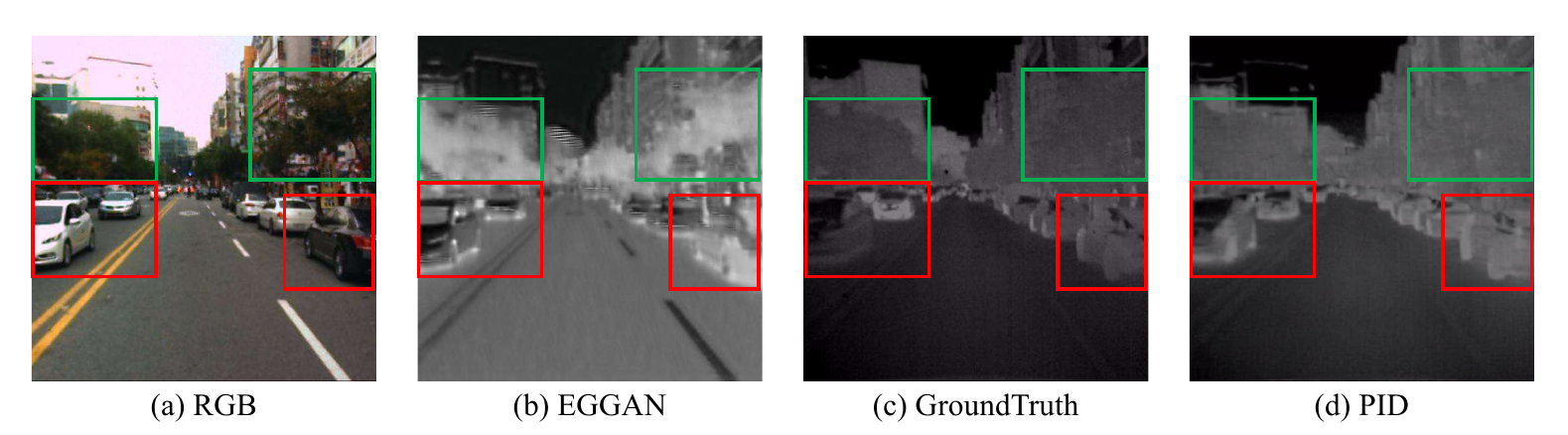}\\
        \caption{Translated infrared image based on RGB image via different methods. The image generated by EGGAN~\cite{Lee2023EdgeguidedMR} shows physical inaccuracies, such as trees appearing hotter than cars (lighter grey color represents higher temperature). Our result is closer to the GroundTruth and adds details in accordance with physical laws, such as the heat generated by moving tires.}
        \label{fig:brief}
\end{figure*}

\textbf{Low generation quality.} Existing methods have utilized Variational Autoencoders~\cite{kingma2013auto} (VAEs) to translate images. Although VAEs are well-suited for image translation, the resulting images frequently lack sharpness~\cite{takida2022preventing}, which makes it challenging for downstream tasks to extract effective infrared features. Generative Adversarial Networks~\cite{Goodfellow2014GenerativeAN} (GANs) have also become a popular approach for generating. However, GAN-based methods often suffer from training instabilities~\cite{Li2022BBDMIT}.

\textbf{Style generation rather than physical generation.} The aforementioned generative works neglect the physical infrared laws. While the generated images may retain edge information, they may differ from the actual physical characteristics of infrared scenes (e.g., a moving cold car under a hot tree generated by EGGAN~\cite{Lee2023EdgeguidedMR} in \autoref{fig:brief}). Such deviations from real-world physical laws may cause accumulated biases in downstream tasks, leading to incorrect recognition of infrared features. 

Over the last few years, Diffusion Models~\cite{Ho2020DenoisingDP,Song2020ScoreBasedGM} (DMs) have gained pronounced progress in image synthesis~\cite{Saharia2021ImageSV,Li2022BBDMIT}. DMs enable to model the rich distribution of the target image domain. DMs demonstrate greater training stability compared to GANs and have achieved exceptional generation quality. By combining the iterative optimization and sampling in the latent space, Latent Diffusion Model~\cite{Rombach2021HighResolutionIS} (LDM) offers an efficient solution for generating high-quality infrared images. These achievements have motivated our exploration of the diffusion-based method for infrared image generation. 

To maintain physical consistency in the generated images, we propose to integrate the principles governing infrared imaging into LDM. The TeX method~\cite{Bao2023HeatassistedDA} decomposes infrared images into fundamental physical components (Temperature (\(T\)), Emissivity (\(e\)), and Thermal texture (\(X\))). It effectively analyzes images in physical space but is hindered by the complexity of the input requirements. Considering that publicly available datasets primarily consist of infrared images directly captured by infrared cameras~\cite{Hwang2015MultispectralPD}, we propose a TeV decomposition method designed for general use. To achieve TeV decomposition, we develop TeVNet ($\mathcal{N}_{\text{TeV}}$) with self-supervised training.

In this paper, we propose PID, a \textbf{P}hysics-\textbf{I}nformed \textbf{D}iffusion model for infrared image generation, which translates RGB images into physically consistent infrared images. To ensure physical consistency, we integrate two physics-based loss functions, $\mathcal{L}_{\text{Rec}}$ and $\mathcal{L}_{\text{TeV}}$, derived from the $\mathcal{N}_{\text{TeV}}$ framework, into the training of LDM. The effectiveness of PID is demonstrated in \autoref{fig:brief}. Our method leverages prior knowledge of infrared physical principles to enforce physical constraints during the training of LDM, thereby improving the fidelity of the generated infrared images to the target domain. In summary, our contributions are as follows:
\begin{itemize}
    \item By analyzing the intrinsic characteristics of infrared image translation, we propose a novel approach that adopts LDM to translate RGB images into high-quality infrared images.
    \item By incorporating physics-based constraints, our method ensures that the translated images adhere to fundamental physical laws, enhancing their applicability in downstream tasks.
    \item We propose an efficient TeV decomposition method for general infrared images captured by infrared cameras, simplifying the requirements of the TeX framework and broadening its practical applicability.
    \item Our proposed method achieves state-of-the-art (\textbf{SOTA}) performance in visible-to-infrared image translation, outperforming GAN-based methods on multiple metrics and datasets.
\end{itemize}

The structure of this paper is as follows: Section \ref{sec:related_work} provides a review in related research areas. Section \ref{sec:methodology} introduces the proposed method in detail. Section \ref{sec:experiments} presents the experiments and their results. Section \ref{sec:discussion} discusses the limitation of proposed method. Section \ref{sec:conclusion} concludes the paper.

\section{Related work}
\label{sec:related_work}

Numerous studies~\cite{Kniaz2018ThermalGANMC, zkanoglu2022InfraGANAG, Lee2023EdgeguidedMR} have focused on image translation methods, enabling the reuse of semantic labels from visible RGB datasets. In the following, we introduce generative models for infrared image translation.

\subsection{Generative models for infrared image translation}  
Variational Autoencoders~\cite{kingma2013auto} (VAEs) learn to encode source RGB images into latent vectors and decode target domain images from sampled latent vectors. However, the simplistic posterior distribution of VAEs often leads to suboptimal generation results.
Generative Adversarial Networks~\cite{Goodfellow2014GenerativeAN} (GANs) improve image translation quality by formulating the generation process as an adversarial game. Conditional GANs~\cite{Mirza2014ConditionalGA} (CGANs) utilize convolutional neural networks (CNNs) to learn mappings from conditional input images to output images. Pix2Pix~\cite{Isola2016ImagetoImageTW}, built upon CGANs, combines the GAN objective with an L1 loss to ensure that translated images closely resemble the ground truth.  
CycleGAN~\cite{Zhu2017UnpairedIT} introduces cycle consistency loss for unpaired image domain transfer tasks. However, as noted by~\cite{Zhang2018SyntheticDG}, unsupervised CycleGAN models are prone to generating unpredictable content during translation.  
To address these limitations, numerous GAN-based approaches~\cite{An2019GeneratingII,Devaguptapu2019BorrowFA,9894231,Yi2023CycleGA} have been proposed, improving network architectures and introducing stronger constraint loss functions to enhance model performance. For instance, Edge-Guided GAN~\cite{Lee2023EdgeguidedMR} emphasizes edge consistency across different image domains. Despite these advancements, GANs remain prone to training instability and often produce translated images of unsatisfactory quality.  

\textit{Our method is based on the latent diffusion model, which can be efficiently trained to convergence and is capable of generating detailed, realistic, and high-quality infrared images from RGB inputs.}

\subsection{Diffusion models for image translation}

Diffusion Models (DMs)~\cite{Ho2020DenoisingDP, Song2020ScoreBasedGM} have recently gained significant attention for their impressive performance in image generation tasks. DMs operate by modeling a Markov chain that progressively transforms Gaussian noise into a target image distribution. During training, the diffusion process involves gradually adding noise to images, and a denoising network (typically built on a UNet architecture) is trained to predict this noise. In the inference stage, the denoising network iteratively predicts and removes the noise, reconstructing the image from the noisy input through a reverse denoising process. This step-by-step denoising enables diffusion models to generate high-quality, clean images, achieving state-of-the-art results in various image processing tasks, such as image generation~\cite{ho2022cascaded}, image restoration~\cite{Kawar2022DenoisingDR}, and image translation~\cite{Li2022BBDMIT}.

Diffusion models are particularly well-suited for fine-grained image translation tasks due to their iterative refinement process~\cite{Saharia2021ImageSV}, which enables them to better handle modality gaps compared to VAEs or GANs. Additionally, they demonstrate superior distribution coverage compared to GANs~\cite{Dhariwal2021DiffusionMB}. To improve computational efficiency, Latent Diffusion Model~\cite{Rombach2021HighResolutionIS} (LDM) encodes images into low-dimensional latent space vectors, performs the diffusion and denoising processes in this latent space, and then decodes the latent vectors back into pixel space to generate high-quality images. This approach significantly reduces computational costs by converting high-dimensional pixel space generation into low-dimensional latent space generation.

Despite the significant progress in diffusion models for RGB image generation, their application to infrared image generation remains underexplored. The infrared image domain differs substantially from the RGB domain, rendering existing pertained RGB generation models ineffective. Furthermore, previous works have primarily treated infrared images as a stylistic variation, neglecting the underlying physical principles that govern infrared imaging.

\textit{Our method addresses these limitations by fully leveraging prior knowledge of infrared imaging to incorporate strong physical constraints during the training of LDM. To the best of our knowledge, we are the first to integrate physical loss into generative models for infrared image translation. Our approach enhances the fidelity of translated images to the target domain without introducing additional inference parameters.}

\section{Methodology}
\label{sec:methodology}

\subsection{Preliminary}
\subsubsection{Diffusion model}  
Given a target domain image \(\boldsymbol{x}_0 \sim p_{\text{data}}\), a series of noisy images \(\{\boldsymbol{x}_1, \boldsymbol{x}_2, \dots, \boldsymbol{x}_T\}\) can be constructed by incrementally adding Gaussian noise. This process is modeled as a Markov chain:  
\begin{align}
    q\left( \boldsymbol{x}_{1:T}|\boldsymbol{x}_{0}\right) &= \prod\nolimits^{T}_{t=1} q\left( \boldsymbol{x}_{t}|\boldsymbol{x}_{t-1}\right), \label{eq1} \\
    \boldsymbol{x}_{t} &= \sqrt{\alpha_{t}} \boldsymbol{x}_{t-1} + \sqrt{1-\alpha_{t}} \boldsymbol{\epsilon}, \label{eq2} \\
    q\left( \boldsymbol{x}_{t}|\boldsymbol{x}_{t-1}\right) &= \mathcal{N}\left( \boldsymbol{x}_{t}|\sqrt{\alpha_{t}}\boldsymbol{x}_{t-1}, \left( 1-\alpha_{t} \right) \mathbf{I} \right), \label{eq3}
\end{align}  
where \(\boldsymbol{\epsilon} \sim \mathcal{N}\left(\mathbf{0}, \mathbf{I} \right)\) is Gaussian noise, and \(\alpha_t \in \left(0,1 \right)\) controls the noise schedule.  

The distribution of \(\boldsymbol{x}_t\) conditioned on \(\boldsymbol{x}_0\) can be derived by simplifying the intermediate steps:  
\begin{align}
    \boldsymbol{x}_{t} &= \sqrt{\bar{\alpha_{t}}} \boldsymbol{x}_{0} + \sqrt{1-\bar{\alpha_{t}}} \boldsymbol{\epsilon}, \label{eq4} \\
    q\left( \boldsymbol{x}_{t}|\boldsymbol{x}_{0}\right) &= \mathcal{N}\left( \boldsymbol{x}_{t}|\sqrt{\bar{\alpha_{t}}} \boldsymbol{x}_{0}, \left( 1-\bar{\alpha_{t}} \right) \mathbf{I} \right), \label{eq5}
\end{align}  
where \(\bar{\alpha_{t}} = \prod\nolimits^{t}_{i=1} \alpha_{i}\) represents the cumulative product of the noise schedule.  

The reverse process also forms a Markov chain, starting from pure Gaussian noise \(\boldsymbol{x}_{T} \sim \mathcal{N}\left( \mathbf{0}, \mathbf{I} \right)\). Through iterative reverse sampling, a high-quality image without noise is obtained:  
\begin{align}
    p_{\theta}\left( \boldsymbol{x}_{0:T}\right) &= p\left( \boldsymbol{x}_{T}\right) \prod\nolimits^{T}_{t=1} p_{\theta}\left( \boldsymbol{x}_{t-1}|\boldsymbol{x}_{t}\right), \label{eq6} \\
    p_{\theta}\left( \boldsymbol{x}_{t-1}|\boldsymbol{x}_{t}\right) &= \mathcal{N}\left( \boldsymbol{\mu}_{\theta} \left(\boldsymbol{x}_{t}; t, \mathbf{c} \right), \sigma^{2}_{t} \mathbf{I} \right). \label{eq7}
\end{align}  
Here, \(\mathbf{c}\) represents a condition, enabling the formulation of a conditional diffusion model.  

The core of the diffusion model lies in training a denoising model \(\boldsymbol{\epsilon}_\theta\) to predict the added noise:  
\begin{equation}
    \theta^{\ast} = \mathop{\arg\min}\limits_{\theta} \mathbb{E}_{\boldsymbol{x} \sim p_{\text{data}}, \boldsymbol{\epsilon} \sim \mathcal{N}(\mathbf{0}, \mathbf{I})}\left[\mathcal{L}_{\text{Noise}}(\boldsymbol{\epsilon}, \boldsymbol{\epsilon}_{\theta} \left( \boldsymbol{x_{t}}; t, \mathbf{c}\right) )\right]. \label{eq8}
\end{equation}  
The predicted noise \(\boldsymbol{\epsilon}_\theta\) is used to compute \(\boldsymbol{\mu_\theta}\), and \(\mathcal{L}_{\text{Noise}}\) is defined as the L1 loss to measure the discrepancy between the predicted noise \(\boldsymbol{\epsilon}_{\theta}\) and the actual noise \(\boldsymbol{\epsilon}\). By focusing on learning independent and identically distributed (i.i.d.) Gaussian noise, the denoising UNet simplifies the modeling of the complex target data distribution. This approach, as demonstrated in DDPM~\cite{Ho2020DenoisingDP}, has been shown to achieve superior sample quality compared to directly predicting the original data \(\boldsymbol{x}_0\). For a complete derivation, please refer to \ref{app1}.

\subsubsection{Latent diffusion model}
Given a target domain image $\boldsymbol{x}_0\in \mathbb{R}^{H\times W \times C}$, LDM~\cite{Rombach2021HighResolutionIS} uses a encoder $\mathcal{E}$ to encode $\boldsymbol{x}_0$ into the latent space vector \(\boldsymbol{z}_0 = \mathcal{E}(\boldsymbol{x}_0) \in \mathbb{R}^{h \times w \times c}\), and the decoder $\mathcal{D}$ reconstructs the image from the latent space vector as $\tilde{\boldsymbol{x}_0}=\mathcal{D}(\boldsymbol{z}_0) = \mathcal{D}(\mathcal{E}(\boldsymbol{x}_0))$. The encoder $\mathcal{E}$ downsamples the image by a factor $f=H/h=W/w$. The diffusion and denoising processes are performed in the latent space. During the diffusion process, the intermediate noisy vectors are generated similarly to the pixel space as \(\boldsymbol{z}_t = \sqrt{\bar{\alpha_t}}\boldsymbol{z}_0 + \sqrt{1-\bar{\alpha_t}}\boldsymbol{\epsilon}\). During the denosing process, the decoder $\mathcal{D}$ translates the results of the denoised latent space vectors back into images in the pixel space. By compressing the sampling space, LDM significantly enhances generation efficiency.

\subsection{TeV decomposition}

\subsubsection{Infrared images decomposition}

TeS~\cite{Gillespie1998ATA} decomposes remote sensing infrared images into temperature $T$ and emissivity $e$. More recently, TeX~\cite{Bao2023HeatassistedDA} uses a neural network to obtain \(T\), \(e\), and thermal texture \(X\). Inspired by these works, we propose to decompose components of the translated infrared images to ensure compliance with physical laws.

\begin{figure}[th]
    \centering
    \includegraphics[width=1\linewidth]{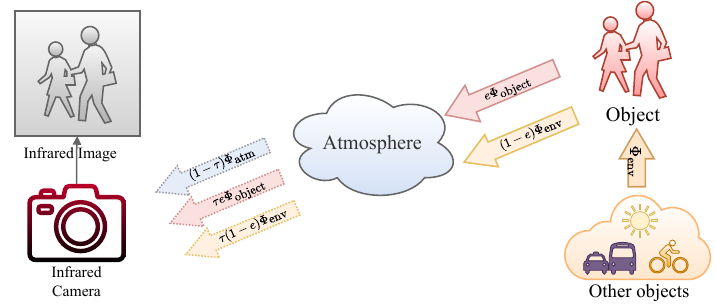}
    \caption{The infrared signal transmission chain when an infrared camera captures an image of pedestrians. The captured infrared signal primarily includes three components~\cite{Vollmer2010InfraredTI}: thermal radiation emitted by the object \(\Phi_{\text{object}}\), thermal radiation reflected from other objects \(\Phi_{\text{env}}\), and atmospheric thermal radiation \(\Phi_{\text{atm}}\). \(\tau\) represents transmissivity of atmosphere while \(e\) represents emissivity of detected object.}
    \label{fig:infrared-detector}
\end{figure}

As illustrated in \autoref{fig:infrared-detector}, a typical infrared image collected by an infrared camera can be described as a superposition of the thermal radiation emitted by the object itself, the reflection of radiation from other objects, and atmospheric radiation.
\begin{equation}
    S_{\lambda}=\tau_{\text{atm}} e_{\lambda} \Phi_{\text{object}} +\tau_{\text{atm}} \left( 1-e_{\lambda}\right)  \Phi_{\text{env}}  \nonumber +\left( 1-\tau_{\text{atm}} \right)  \Phi_{\text{atm}}.
\end{equation}
\(S_{\lambda}\) represents the thermal radiation intensity at detected wavelength \(\lambda\). The first term represents the thermal radiation from object itself. It can be expressed as 
\begin{equation}
\tau_{\text{atm}} e_{ \lambda}\Phi_{\text{object}} = \tau_{\text{atm}} e_{ \lambda} B_{\lambda}\left( T\right),
\end{equation}
where \(\tau_{\text{atm}}\) represents the transmissivity of atmosphere~\cite{Incropera2018PrinciplesOH}, and \(e_{\lambda}\) denotes the object's emissivity. \(e_{\lambda}\) is only related to the material of the object. A lower \(e_{\lambda}\) indicates a better radiation reflection ability. The spectral irradiance of a heating body \(\Phi_{\text{object}}\) is determined by Planck's black body law~\cite{planck1900theory} with \(B_{\lambda}\left( T\right)\). 
\begin{equation}
    B_{\lambda}(T) =\frac{2\pi hc^{2}}{\lambda^{5}} \frac{1}{e^{\frac{hc}{\lambda kT}}-1},
\end{equation}
where \(h\) is the Planck constant, \(k\) denotes the Boltzmann constant, and \(c\) represents the speed of light. \(T\) stands for the temperature of the object. The reflection of radiation from other objects in the environment and atmospheric radiation can be summarized as, 
\begin{equation}
    \tau_{\text{atm}} \left( 1-e_{\lambda}\right)  \Phi_{\text{env}}  +\left( 1-\tau_{\text{atm}} \right)  \Phi_{\text{atm}}, 
\end{equation}
where \(\Phi_{\text{env}}\) and \(\Phi_{\text{atm}}\) denote the radiation from other objects and atmosphere, respectively. 
Due to the absorption of infrared radiation in the atmosphere, captured infrared images typically have \(\tau_{\text{atm}} \approx 1\), leading to the approximation of $S_{\lambda}$, 
\begin{equation}
    S_{\lambda} \approx e_{ \lambda}B_{\lambda}\left( T\right)  + \left( 1-e_{\lambda}\right)  \Phi_{\text{env}}.
\end{equation}

TeX~\cite{Bao2023HeatassistedDA} proposes that \(\Phi_{\text{env}}\) at pixel coordinates $\alpha=(x, y)$ originates from a linear combination of the infrared signal values of its neighboring pixels $\beta=\{(x_1, y_1), ..., (x_{N-1}, y_{N-1})\}$, where $N = H \times W$ denotes the total number of pixels in the infrared image.
\begin{equation}
    \Phi_{\text{env},\alpha} \approx X_{\lambda \alpha}=\sum_{i=1}^{N-1} V_{\alpha \beta_i}S_{\lambda \beta_i}.
\end{equation}
Thus, TeX decomposes the $S_{\lambda \alpha}$ at $\alpha=(x, y)$ as 
\begin{align}
    S_{\lambda \alpha}=e_{\lambda \alpha}B_{\lambda}\left( T_{\alpha}\right)  +\left( 1-e_{\lambda \alpha}\right)  X_{\lambda \alpha},
    \label{eq:tex}
\end{align}
where $V_{\alpha \beta_i}$ represents the linear combination vector from pixel $\beta_i$ towards pixel $\alpha$. Due to the high computational complexity of $\mathcal{O}(N^2)$ required to predict $V_{\alpha \beta_i}$ between every pair of pixels, TeX propose a more efficient approach. By dividing the infrared image into $m$ grids and performing local averaging, the full matrix $\mathbf{S}_\lambda \in \mathbb{R}^{H \times W}$ is downsampled to $\hat{\mathbf{S}}_{\lambda} \in \mathbb{R}^{m \times 1}$. The components of $\hat{\mathbf{S}}_{\lambda}$ are then linearly combined with the corresponding coefficients $\mathbf{V}_{\lambda \alpha} \in \mathbb{R}^{1 \times m}$ to approximate $\Phi_{\text{env},\alpha}$. As shown in Eq.~\eqref{eq:VS}, this approach reduces the computational complexity to $\mathcal{O}(N)$.
\begin{equation}
    \label{eq:VS}
    X_{\lambda \alpha} \approx \mathbf{V}_{\lambda \alpha} \hat{\mathbf{S}}_{\lambda},\ 
    \hat{\mathbf{S}}_{\lambda} = \text{Ds}(\mathbf{S}_{\lambda}).
\end{equation}

\subsubsection{TeV decomposition for spectral integration}
\label{sec:Spectral integration}

According to \cite{Bao2023HeatassistedDA}, the TeX method requires infrared signals at different wavelengths. However, most currently available public datasets consist of infrared images captured via spectral integration by cameras. This discrepancy creates a gap between the TeX method and general infrared images, making the TeX method unsuitable for infrared generative models. 
To address this gap, we propose the TeV decomposition method. In the following, we will introduce our TeV decomposition method in detail.

Based on Eq.~\eqautoref{eq:tex} and Eq.~\eqautoref{eq:VS}, we propose Eq.~\eqautoref{eq:tev-1}, which represents the superimposition of multiple infrared spectral signals. \([\lambda_{min},\lambda_{max}]\) is the working wavelength range for an infrared camera.
\begin{align}
    \int_{\lambda_{min}}^{\lambda_{max}} S_{\lambda \alpha}d\lambda =\int_{\lambda_{min}}^{\lambda_{max}} e_{\lambda \alpha}B_{\lambda}\left( T_{\alpha}\right)  d\lambda +\int_{\lambda_{min}}^{\lambda_{max}} \left( 1-e_{\lambda \alpha}\right)\mathbf{V}_{\lambda \alpha} \hat{\mathbf{S}}_{\lambda} d\lambda .
    \label{eq:tev-1}
\end{align}

Available data in \cite{Bao2023HeatassistedDA} indicates that, the emissivity $e_{\lambda}$ of most objects in common scenarios remains relatively constant across the working wavelengths. The emissivities of common materials are shown in \ref{app2}. Therefore, \textbf{we assume a constant emissivity for a specific material.} 
\begin{align}
    \mathcal{S}_{\alpha} =e_{\alpha} \int_{\lambda_{min}}^{\lambda_{max}} B_{\lambda}\left( T_{\alpha}\right)  d\lambda +\left( 1-e_{\alpha}\right)  \int_{\lambda_{min}}^{\lambda_{max}} \mathbf{V}_{\lambda \alpha} \hat{\mathbf{S}}_{\lambda} d\lambda .
\end{align}
We define the first integration as \(\mathbf{T}_{\alpha}\), which represents the radiation emitted by the object itself at pixel \(\alpha\). 
\begin{align}
\mathbf{T}_{\alpha}=\int_{\lambda_{min}}^{\lambda_{max}} B_{\lambda}\left( T_{\alpha}\right)  d\lambda .
\end{align}
The second integration can also be simplified as a product of two matrices, which represents the reflective of other objects' radiation. 
\begin{equation}
    \int_{\lambda_{min}}^{\lambda_{max}} \mathbf{V}_{\lambda \alpha} \hat{\mathbf{S}}_{\lambda} d\lambda =\underbrace{\mu_{\alpha} \int_{\lambda_{min}}^{\lambda_{max}} \mathbf{V}_{\lambda \alpha}d\lambda}_{\mathbf{V}_{\alpha}}   \underbrace{\int_{\lambda_{min}}^{\lambda_{max}} \hat{\mathbf{S}}_{\lambda} d\lambda \nonumber}_{\hat{\mathcal{S}}} =\mathbf{V}_{\alpha} \hat{\mathcal{S}} .
\end{equation}
Here, \(\hat{\mathcal{S}}\) represents the average downsampling result of the complete infrared image \(\mathcal{S}\), corresponding to the operation described in Eq.~\eqautoref{eq:VS}.
\begin{align}
    \int_{\lambda_{min}}^{\lambda_{max}} \hat{\mathbf{S}}_{\lambda} d\lambda =\int_{\lambda_{min}}^{\lambda_{max}} \text{Ds} \left( \mathbf{S}_{\lambda}\right)  d\lambda =\text{Ds} (\int_{\lambda_{min}}^{\lambda_{max}} \mathbf{S}_{\lambda}d\lambda )=\text{Ds}(\mathcal{S})=\hat{\mathcal{S}} .
\end{align}
Thus, we rewrite the decomposition equation for the infrared spectral integration, named as the \textbf{TeV decomposition method}.
\begin{equation}
    \label{eq:rec}
    \mathcal{S} =\mathbf{e} \odot \mathbf{T} +\left(\mathbf{1}-\mathbf{e}\right) \odot \mathbf{V} \hat{\mathcal{S}},
\end{equation}
where $\odot$ represents element-wise multiplication of matrices, \(\mathbf{e}=\{e_\alpha|\alpha=(x,y)\} \in \mathbb{R}^{H \times W} \) represents the emissivity matrix, \(\mathbf{T} =\{\mathbf{T}_\alpha|\alpha=(x,y)\}\in \mathbb{R}^{H \times W} \) represents the temperature matrix, \(\mathbf{V}=\{\mathbf{V}_\alpha|\alpha=(x,y)\} \in \mathbb{R}^{H \times W \times m} \) represents the thermal vector matrix. 

\begin{figure*}[!htbp]
    \centering
    \includegraphics[width=0.9\linewidth]{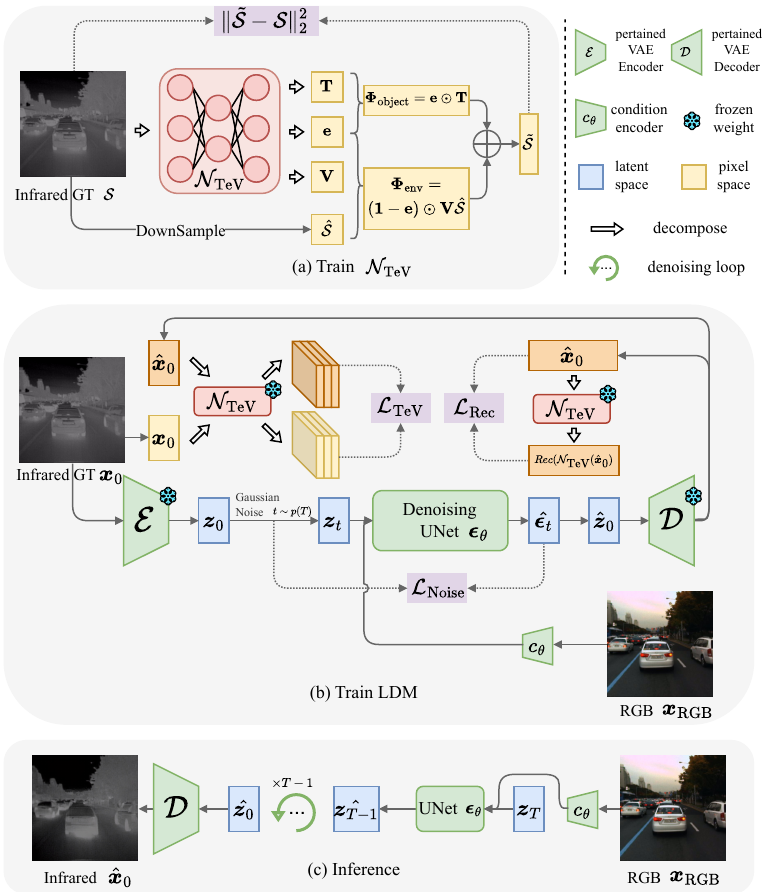}
    \caption{The overview of our proposed PID. (a) Pretraining of $\mathcal{N}_{\text{TeV}}$: the PID trains a $\mathcal{N}_{\text{TeV}}$ with self-supervised loss on infrared dataset. (b) Physics-informed latent diffusion model training: the infrared GT image $\boldsymbol{x}_0$ is encoded by the pretrained encoder $\mathcal{E}$ to obtain the latent space vector \(\boldsymbol{z}_0\). Gaussian noise is then added to \(\boldsymbol{z}_0\) to produce intermediate vector \(\boldsymbol{z}_t\). The denoising UNet is trained to predict the added noise, with the RGB image condition. In addition to \(\mathcal{L}_{\text{Noise}}\), $\mathcal{L}_{\text{TeV}}$ and $\mathcal{L}_{\text{Rec}}$ are also incorporated into the training process. During this process, the weights of $\mathcal{E}$, $\mathcal{D}$ and $\mathcal{N}_{\text{TeV}}$ are frozen, allowing PID to learn the infrared features and physical laws without increasing extra training parameters. (c) The inference process of PID.}
    \label{fig:overview}
\end{figure*}

Based on the derived relationship, as shown in \autoref{fig:overview}(a), we can train a self-supervised decomposition network $\mathcal{N}_{\text{TeV}}$ to self-adaptively predict \(\mathbf{e}\), \(\mathbf{T}\), and \(\mathbf{V}\) by optimizing $\mathbb{E}_{\mathcal{S} \in p_{\text{data}}} \| \tilde{\mathcal{S}} -\mathcal{S} \|_2^2$
\begin{equation}
    \label{eq:dec}
    \tilde{\mathbf{e}} ,\tilde{\mathbf{T}} ,\tilde{\mathbf{V}}=\mathcal{N}_{\text{TeV}} \left( \mathcal{S} \right),
\end{equation}
where $\mathcal{S}$ is the ground truth and $\tilde{\mathcal{S}}$ is the reconstructed infrared image. Rec represents the reconstruction method described in Eq.~\eqautoref{eq:rec}.
\begin{equation}
    \tilde{\mathcal{S}} = \text{Rec}(\tilde{\mathbf{e}} ,\tilde{\mathbf{T}} ,\tilde{\mathbf{V}}).
\end{equation}

The TeV decomposition method addresses the limitations of TeX by leveraging the material-specific emissivity assumption. This allows the TeV decomposition method to handle general infrared images without requiring complex input, making it more practical for addressing the physical constraints of infrared diffusion models.

\subsubsection{Physics-informed losses}

The translated infrared images from current works may exhibit violations of physical laws. To address this issue, we introduce two physics-informed losses to supervise the diffusion model training, ensuring that the generated images are physically meaningful and consistent with the real world.

\noindent\textbf{Physical reconstruction loss $\mathcal{L}_{\text{Rec}}$}. The first loss, $\mathcal{L}_{\text{Rec}}$, is designed to enforce the physical consistency of the translated image. As shown in Eq.~\eqautoref{eq:lrec}, $\mathcal{L}_\text{MSE}$ is the mean square error loss. It measures the discrepancy between the translated image and the infrared domain. By minimizing this loss, the diffusion model is encouraged to generate images that not only appear realistic but also adhere to the physical principles governing the infrared domain. \(\mathcal{L}_{\text{Rec}}\) acts like a discriminator in GANs but operates continuously in the diffusion model.
\begin{equation}
    \label{eq:lrec}
    \mathcal{L}_{\text{Rec}} = \mathcal{L}_\text{MSE}(\text{Rec} (\mathcal{N}_{\text{TeV}} (\hat{\boldsymbol{x}}_{0} )), \hat{\boldsymbol{x}}_{0}).
\end{equation}

\noindent\textbf{Physical TeV space loss $\mathcal{L}_{\text{TeV}}$}. The second loss, $\mathcal{L}_{\text{TeV}}$, operates in the TeV space to enforce physical consistency between the translated and ground truth images. This loss ensures that the translated images are not only visually similar to the ground truth but also physically similarly in the TeV space.
\begin{equation}
    \mathcal{L}_{\text{TeV}} = \mathcal{L}_\text{MSE}(\mathcal{N}_{\text{TeV}} (\hat{\boldsymbol{x}}_{0} ), \mathcal{N}_{\text{TeV}} (\boldsymbol{x}_{0})).
\end{equation}

The combination of $\mathcal{L}_{\text{Rec}}$ and $\mathcal{L}_{\text{TeV}}$ ensures that the model not only learns to generate realistic images but also understands the physical principles that underlie the infrared domain.

\subsection{PID architecture}
Combining efficient physical losses with the LDM, we propose a PID architecture, as illustrated in \autoref{fig:overview}. The framework comprises three main stages: TeVNet training, LDM training, and LDM inference.

\subsubsection{TeVNet training}
In the TeVNet training stage (see \autoref{fig:overview}(a)), we first train \(\mathcal{N}_{\text{TeV}}\) using the self-supervised method described in Section \ref{sec:Spectral integration}. Once trained, the weights of \(\mathcal{N}_{\text{TeV}}\) are frozen and remain fixed throughout subsequent stages.

\subsubsection{LDM training}
In the LDM training stage (see \autoref{fig:overview}(b)), we encode the ground truth infrared image \(\boldsymbol{x}_{0}\) into a latent space vector \(\boldsymbol{z}_0 = \mathcal{E}(\boldsymbol{x}_{0})\). At the same time, the RGB image \(\boldsymbol{x}_{\text{RGB}}\) is encoded by a conditional encoder to produce \(\mathbf{c}_{\text{RGB}} = c_\theta (\boldsymbol{x}_{\text{RGB}})\). Both \(\boldsymbol{z}_0\) and \(\mathbf{c}_{\text{RGB}}\) have the same spatial dimension \(h \times w\). We then add Gaussian noise \(\boldsymbol{\epsilon} \sim \mathcal{N}(\mathbf{0}, \mathbf{I})\) to \(\boldsymbol{z}_0\), resulting in \(\boldsymbol{z}_t = \sqrt{\bar{\alpha_t}}\boldsymbol{z}_0 + \sqrt{1-\bar{\alpha_t}}\boldsymbol{\epsilon}\). The conditional image \(\mathbf{c}_{\text{RGB}}\) is concatenated with \(\boldsymbol{z}_t\) along the channel dimension, and we train a denoising model \(\boldsymbol{\epsilon}_\theta(\boldsymbol{z}_t; t, \mathbf{c}_{\text{RGB}})\) to predict the noise \(\boldsymbol{\epsilon}\).

This stage introduces two key innovations: the \textbf{physical reconstruction loss} \(\mathcal{L}_{\text{Rec}}\) and the \textbf{TeV space loss} \(\mathcal{L}_{\text{TeV}}\). Both losses are computed using the predicted \(\hat{\boldsymbol{x}_{0}}\), which is derived as:
\begin{equation}
    \hat{\boldsymbol{x}_{0}} = \mathcal{D} \left( \frac{\boldsymbol{z}_{t} - \sqrt{1-\bar{\alpha_{t}}} \boldsymbol{\epsilon}_{\theta}}{\sqrt{\bar{\alpha_{t}}}} \right),
\end{equation}
where \(\boldsymbol{\epsilon}_{\theta}\) is the predicted noise, and \(\mathcal{D}\) is the decoder. The pertained model \(\mathcal{N}_{\text{TeV}}\) is utilized to compute both \(\mathcal{L}_{\text{Rec}}\) and \(\mathcal{L}_{\text{TeV}}\). These losses are integrated into the training objective alongside the \(\mathcal{L}_{\text{Noise}}(\boldsymbol{\epsilon}, \boldsymbol{\epsilon}_{\theta})\), scaled by hyperparameters \(k_1\) and \(k_2\). Detailed pseudo-code can be referred to \ref{app3} \autoref{al:tevnet}. The overall optimization objective is formulated as:
\begin{align}
\theta^{\ast}_{PID} = \mathop{\arg\min}\limits_{\theta}\ \mathbb{E}_{\boldsymbol{x}_{0} \sim p_{data}, \boldsymbol{\epsilon} \sim \mathcal{N}(\mathbf{0}, \mathbf{I})} \notag \\
\mathcal{L}_{\text{Noise}}(\boldsymbol{\epsilon}, \boldsymbol{\epsilon}_{\theta} \left( \boldsymbol{z}_{t}; t, \mathbf{c}_{\text{RGB}}\right))
+ k_1\mathcal{L}_{\text{Rec}} + k_2\mathcal{L}_{\text{TeV}}.
\label{eq:theta_pid}
\end{align}

\subsubsection{LDM inference}

The inference stage (see \autoref{fig:overview}(c)) begins by sampling Gaussian noise in the latent space at \(t=T\). The model iteratively removes noise until \(t=0\), and the decoder \(\mathcal{D}\) reconstructs the final infrared image \(\hat{\boldsymbol{x}_0} = \mathcal{D}(\hat{\boldsymbol{z}}_0)\), where \(\hat{\boldsymbol{z}}_0\) is the result of denoised latent space vector.

\section{Experiments results and analysis}
\label{sec:experiments}

\subsection{Experiments setup}

\subsubsection{Datasets}
According to Wien's displacement law~\cite{Griffiths2017IntroductionTE}, the wavelength of infrared images with the strongest energy at \(20^{\circ}\mathrm{C}\) (293.15 K) can be calculated as:
\begin{equation}
    \lambda_{\text{max}} = \frac{b}{\text{Temperature}} = \frac{2898\ \mu m \cdot K}{293.15\ K} \approx 10\ \mu m,
\end{equation}
where \(b\) is Wien's displacement constant. In real-world applications, the wavelength of infrared images typically falls within the long-wavelength band. Therefore, we select the KAIST~\cite{Hwang2015MultispectralPD} (7.5–13.5 \(\mu m\)) and FLIR~\cite{Flirdataset} (7.5–13.5 \(\mu m\)) datasets, both of which lie within this range. To evaluate the generalization of PID, we also utilize the VEDAI~\cite{Razakarivony2016VehicleDI} (700–2500 nm) near-infrared dataset. For the KAIST dataset, we use 12,538 image pairs for training and 2,252 for testing. For the FLIR dataset, we use 8,347 image pairs for training and 1,256 for testing. For the VEDAI dataset, we use 1,000 image pairs for training and 268 for testing.

\subsubsection{Data preprocessing}
The KAIST and VEDAI datasets are provided pre-aligned by their authors. For the FLIR dataset, alignment is performed by manually applying resizing and shifting corrections based on matching. All images are cropped into centered squares based on their shortest side length and resized to \(512 \times 512\). Pixel values are normalized to the range \([-1, 1]\). To enhance model generalization, random cropping and flipping are applied as data augmentation techniques during training.

\subsubsection{TeVNet training details}
In the first stage, we train \(\mathcal{N}_{\text{TeV}}\) using a UNet~\cite{Ronneberger2015UNetCN} architecture with a ResNet18~\cite{He2015DeepRL} encoder pretrained on ImageNet~\cite{deng2009imagenet} (sourced from \cite{Iakubovskii:2019}). The number of channels for \(\mathbf{V}\) is set to \(m=4\). To ensure consistency between decomposition values and their physical meanings, a sigmoid activation function is applied to the \(\mathbf{e}\) output layer, and a ReLU activation function is applied to the \(\mathbf{T}\) output layer. The \(\mathcal{N}_{\text{TeV}}\) models are trained using self-supervised loss on the training datasets until convergence. Further architectural and training details are provided in \ref{app3}.

\subsubsection{LDM training details}
In the second stage, the ground truth infrared images are encoded into \(64 \times 64 \times 4\) latent space vectors using a pretrained VQGAN-f8~\cite{Rombach2021HighResolutionIS}. Two types of conditioners are used to encode the visible RGB guide image: a 3-layer MLP $\mathcal{M}$ (trainable) and the latent space encoder \(\mathcal{E}\) (frozen). During training, the weights of \(\mathcal{N}_{\text{TeV}}\) remain frozen. Following previous work~\cite{Rombach2021HighResolutionIS}, we set 1,000 diffusion steps for training and use \(s \le 200\) steps for inference with DDIM~\cite{Song2020DenoisingDI}. Through experiments, the loss weights are set to \(k_1=50\) and \(k_2=5\) for the KAIST dataset, \(k_1=k_2=50\) for the FLIR dataset, and \(k_1=k_2=20\) for the VEDAI dataset. Additional architectural and training details are provided in \ref{app3}.

\subsubsection{Evaluation metrics}
We evaluate performance using commonly used image similarity metrics, including Structural Similarity Index~\cite{Wang2004ImageQA} (SSIM) and Peak Signal-to-Noise Ratio (PSNR). Previous study~\cite{Saharia2021ImageSV} suggests that Learned Perceptual Image Patch Similarity~\cite{Zhang2018TheUE} (LPIPS) and Fréchet Inception Distance~\cite{Heusel2017GANsTB} (FID) better reflect the perceptual quality of generated images, and thus they are also included in our evaluation.

Section \ref{sec:kaist} presents the performance of PID on the KAIST dataset, while Section \ref{sec:flir} demonstrates its performance on the FLIR dataset. Section \ref{sec:vedai} shows the results on the VEDAI dataset. Finally, Section \ref{sec:ablation} provides an ablation study of the proposed method.

\subsection{Experimental results}

\subsubsection{Results on KAIST dataset}

\label{sec:kaist}

\begin{table*}[ht]
    \setlength\tabcolsep{4pt}
    \centering
    \caption{Quantitative results on the KAIST dataset. $\dag$ represents the available data on published papers~\cite{zkanoglu2022InfraGANAG}. The best results are highlighted in \textbf{bold}, and the second-best results are \underline{underlined}.}
    \label{tab:kaist}
    \scriptsize
    \begin{tabular}{lccllllccc}
    \hline
        \textbf{Methods} & \makecell[c]{\textbf{Publication}\\\textbf{Venue}} & \makecell[c]{\textbf{Pre-}\\\textbf{trained}} & \textbf{SSIM}$\uparrow$ & \textbf{PSNR}$\uparrow$ & \textbf{LPIPS}$\downarrow$ & \textbf{FID}$\downarrow$ & \makecell[c]{\textbf{Param.}$\downarrow$\\\textbf{(M)}} & {\makecell[c]{\textbf{MACs}$\downarrow$\\\textbf{(G)}}} & {\textbf{Type}} \\ \hline
        \textbf{Pix2Pix~\cite{Isola2016ImagetoImageTW}} & \text{CVPR 2017} & - & 0.69$^\dag$ & 21.25$^\dag$ & 0.196$^\dag$ & 132.04 & {54.41} & {72.61} & {GAN} \\ 
        \textbf{CycleGAN~\cite{Zhu2017UnpairedIT}} & \text{ICCV 2017} & - & 0.6016$\pm$0.1204 & 17.02$\pm$2.71 &  0.2515$\pm$0.0718&  67.57 & {11.38} & {227.46} & {GAN} \\ 
        \textbf{ThermalGAN~\cite{Kniaz2018ThermalGANMC}} & \text{CVPR 2019} & - & 0.66$^\dag$ & 19.74$^\dag$ & 0.242$^\dag$ & 277.85 & {66.99} & {71.62} & {GAN} \\ 
        {\textbf{PSP~\cite{Richardson2020EncodingIS}}} & {\text{CVPR 2021}} & {IR-SE50~\cite{Deng2018ArcFaceAA}} & {0.7605$\pm$0.0945} & {22.76$\pm$3.67} &  {0.1812$\pm$0.0612} & {176.30} & {234.09} & {66.18} & {GAN} \\
        \textbf{InfraGAN~\cite{zkanoglu2022InfraGANAG}} & \text{PRL 2022} & - & 0.76$^\dag$&  22.97$^\dag$ &  0.159$^\dag$ &  222.96 & {66.99} & {71.62} & {GAN} \\
        \textbf{EGGAN-O~\cite{Lee2023EdgeguidedMR}} & \text{ICRA 2023} & EGGAN~\cite{Lee2023EdgeguidedMR} & 0.4486$\pm$0.1204 &  9.91$\pm$2.43 & 0.3687$\pm$0.0662&  96.83 & {11.14} & {284.35} & {GAN} \\ 
        \textbf{EGGAN-U~\cite{Lee2023EdgeguidedMR}} & \text{ICRA 2023} & - & 0.6260$\pm$0.1078& 17.07$\pm$3.26 & 0.2747$\pm$0.0646 &  76.27 & {15.03} & {309.3} & {GAN} \\ 
        \textbf{EGGAN-M~\cite{Lee2023EdgeguidedMR}} & \text{ICRA 2023} & - & 0.4782$\pm$0.1028 & 10.65$\pm$1.83 & 0.3685$\pm$0.0770 & 79.45 & {11.14} & {284.35} & {GAN} \\ 
        {\textbf{IRFormer~\cite{Chen2024ImplicitMT}}} & {\text{IJCNN 2024}} & {-} & {0.7959$\pm$0.0943} & {22.19$\pm$3.13} &  {0.2507$\pm$0.0938} & {227.52} & {0.039} & {9.66} & {ViT} \\
        {\textbf{StegoGAN~\cite{Wu_2024_CVPR}}} & {\text{CVPR 2024}} & {-} & {0.5911$\pm$0.1077} & {16.15$\pm$2.61} & {0.2628$\pm$0.0529} & {82.50} & {11.37} & {227.46} & {GAN} \\
        {\textbf{CG-turbo~\cite{Parmar2024OneStepIT}}} & {\text{Arxiv 2024}} & {SD-turbo~\cite{Sauer2023AdversarialDD}} & {0.6823$\pm$0.1122} & {19.95$\pm$3.62} &  {0.1947$\pm$0.0641} & {56.95} & {1540.28} & {16942.27} & {GAN} \\ \hline
        \textbf{LDM-f4($s=200, c_\theta=\mathcal{M}$)} & \text{CVPR 2022} & - & 0.6025$\pm$0.1729 & 14.77$\pm$5.79 & 0.1948$\pm$0.0828 & 78.68 & - & - & {DM} \\ 
        \textbf{LDM-f8($s=200, c_\theta=\mathcal{M }$)} & \text{CVPR 2022} & - & 0.7892$\pm$0.0940 & 23.31$\pm$4.00 & 0.1380$\pm$0.0581 & 64.54 & {281.65} & {23813.62} & {DM} \\ 
        \textbf{PID-f8($s=200, c_\theta=\mathcal{M}$)} & \text{-} & - & 0.7913$\pm$0.0931 & 23.60$\pm$3.97 & \underline{0.1366$\pm$0.0582} & 51.69 & {281.65} & {23813.62} & {DM} \\
        {\textbf{PID-f8($s=20, c_\theta=\mathcal{M}$)}} & {\text{-}} & {-} & {\underline{0.7981$\pm$0.0942}} & {\underline{23.65$\pm$4.03}} & {0.1389$\pm$0.0626} & {\underline{45.01}} & {281.65} & {3216.22} & {DM} \\
        {\textbf{PID-f8($s=20, c_\theta=\mathcal{E}$)}} & {-} & {-} & {\textbf{0.8097$\pm$0.0917}} & {\textbf{24.24$\pm$3.91}} & {\textbf{0.1282$\pm$0.0583}} & {\textbf{44.81}} & {{309.18}} & {{3655.56}} & {DM} \\\hline
    \end{tabular}
\end{table*}

\begin{figure*}[!ht]
    \centering 
    \includegraphics[width=1\linewidth]{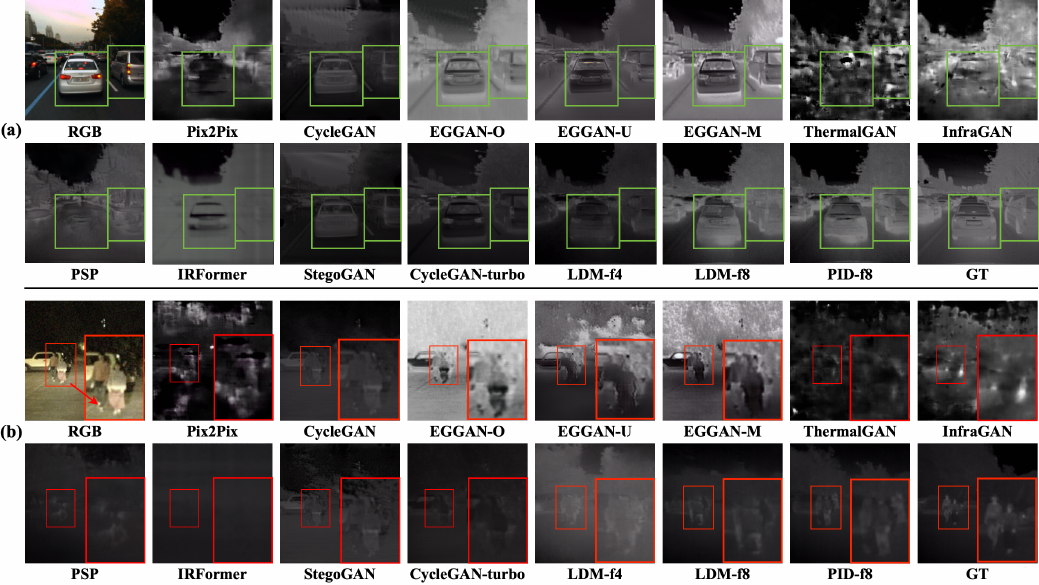}
    \caption{{Qualitative results on KAIST dataset. For the sake of clarity, the magnified region is highlighted with boxes for easier comparison. PID demonstrates strong robustness in both daytime and nighttime scenes.}}
    \label{fig:kaist}
\end{figure*}

The KAIST dataset includes three typical scenes: campus, road, and downtown, each captured during both daytime and nighttime. These scenes feature essential elements for driving, such as roads, vehicles, and pedestrians.

\autoref{tab:kaist} shows the results on the KAIST dataset. We compare not only GAN-based methods but also LDM with various $\mathcal{E}$ and $\mathcal{D}$ configurations. Specifically, f4 and f8 represent the encoder and decoder of VQGAN with $f=4$ and $f=8$, respectively.

Our experiments demonstrate that the diffusion model not only improves structural similarity but also enhances perceptual quality compared to GAN-based methods. Specifically, LDM-f8 outperforms InfraGAN~\cite{zkanoglu2022InfraGANAG} by 0.0292 in SSIM, 0.34 in PSNR, reduces LPIPS by 0.021, and significantly lowers FID by 158.42. Moreover, LDM-f8 outperforms LDM-f4, likely due to the limited capacity of the VQGAN-f4 architecture, which may cause overfitting of the denoising UNet.

Our proposed PID method, built upon LDM and incorporating physical losses, demonstrates superior performance compared to the baseline LDM method and other GAN-based model. When compared to LDM-f8, PID-f8 ($s=20, c_\theta=\mathcal{M}$) achieves improvements of 0.0089 in SSIM, 0.34 in PSNR, and a significant reduction of 19.53 in FID. Notably, PID-f8 ($s=20, c_\theta=\mathcal{E}$) achieves the best overall performance among the configurations tested. Compared to CycleGAN-turbo~\cite{Parmar2024OneStepIT} (CG-turbo), our method not only outperforms it in terms of generative quality but also achieves this with lower computational cost.

Additionally, we observe that excessive sampling may introduce instability, potentially degrading the quality of generated samples. Furthermore, replacing the MLP-based conditioner $c_\theta=\mathcal{M}$ with the encoder $\mathcal{E}$, which facilitates transformations between pixel and latent spaces, results in notable improvements in generative performance. While this replacement introduces only a small increase in inference parameters. The gains arise from the high structural similarity between dual-modality images in the pixel space, with $\mathcal{E}$ enabling their encoding into a shared latent space to improve the diffusion model's ability to capture their spatial correspondence.

\autoref{fig:kaist} shows qualitative results on the KAIST dataset. Our approach generates more accurate infrared semantic information than baseline methods. For example, the road surface beneath the cars in the ground truth (\autoref{fig:kaist}(a)) shows strong signal intensity, which PID and LDM both capture, with PID providing clearer boundaries due to its focus on physical properties.

A key challenge in visible-to-infrared translation is poor illumination in the RGB input. Baseline methods perform poorly in such cases. For instance, in \autoref{fig:kaist}(b), EGGAN~\cite{Lee2023EdgeguidedMR} results show pedestrians cooler than their surroundings, violating physical laws. In contrast, PID captures the infrared properties of pedestrians more effectively, producing clearer images and better handling nighttime conditions.

\subsubsection{Results on FLIR dataset} \label{sec:flir}

\begin{table*}[ht]
    \setlength\tabcolsep{4pt}
    \centering
    \caption{Quantitative metrics on FLIR dataset. The best results are highlighted in \textbf{bold}, and the second-best results are \underline{underlined}.}
    \label{tab:flir}
    \scriptsize
    \begin{tabular}{lccllllccc}
    \hline
        \textbf{Methods} & \makecell[c]{\textbf{Publication}\\\textbf{Venue}} & \makecell[c]{\textbf{Pre-}\\\textbf{trained}} & \textbf{SSIM}$\uparrow$ & \textbf{PSNR}$\uparrow$ & \textbf{LPIPS}$\downarrow$ & \textbf{FID}$\downarrow$ & {\makecell[c]{\textbf{Param.}$\downarrow$\\\textbf{(M)}}} & {\makecell[c]{\textbf{MACs}$\downarrow$\\\textbf{(G)}}} & {\textbf{Type}}  \\ \hline
        \textbf{Pix2Pix~\cite{Isola2016ImagetoImageTW}} & \text{CVPR 2017} & - & 0.2986$\pm$0.0976 & 16.34$\pm$2.57 & 0.3833$\pm$0.0592&  199.06 & {54.41} & {72.61} & {GAN} \\ 
        \textbf{CycleGAN~\cite{Zhu2017UnpairedIT}} & \text{ICCV 2017} & - & 0.3451$\pm$0.0800&  12.74$\pm$1.62&  0.4270$\pm$0.0653&  100.32 & {11.38} & {227.46} & {GAN} \\ 
        \textbf{ThermalGAN~\cite{Kniaz2018ThermalGANMC}} & \text{CVPR 2019} & - & 0.1523$\pm$0.0697&  11.59$\pm$1.11&  0.5981$\pm$0.0492&  407.01 & {66.99} & {71.62} & {GAN} \\ 
        {\textbf{PSP~\cite{Richardson2020EncodingIS}}} & {\text{CVPR 2021}} & {IR-SE50~\cite{Deng2018ArcFaceAA}} & {0.3829$\pm$0.1007} & {17.70$\pm$1.84} &  {0.3889$\pm$0.0568} & {301.30} & {234.09} & {66.18} & {GAN} \\
        \textbf{InfraGAN~\cite{zkanoglu2022InfraGANAG}} & \text{PRL 2022} & - & 0.4050$\pm$0.1138&  16.95$\pm$2.11&  0.5050$\pm$0.0847&  399.71 & {66.99} & {71.62} & {GAN} \\
        \textbf{EGGAN-O~\cite{Lee2023EdgeguidedMR}} & \text{ICRA 2023} & EGGAN~\cite{Lee2023EdgeguidedMR} & 0.3464$\pm$0.0839&  9.65$\pm$0.8121&  0.5564$\pm$0.0731&  140.01 & {11.14} & {284.35} & {GAN} \\ 
        \textbf{EGGAN-U~\cite{Lee2023EdgeguidedMR}} & \text{ICRA 2023} & - & 0.3745$\pm$0.1009&  14.30$\pm$2.19&  0.4719$\pm$0.0815&  140.92 & {15.03} & {309.3} & {GAN} \\ 
        \textbf{EGGAN-M~\cite{Lee2023EdgeguidedMR}} & \text{ICRA 2023} & - & 0.2904$\pm$0.0686&  10.32$\pm$1.42&  0.4345$\pm$0.0496&  139.11 & {11.14} & {284.35} & {GAN} \\ 
        {\textbf{IRFormer~\cite{Chen2024ImplicitMT}}} & {\text{IJCNN 2024}} & {-} & {\textbf{0.4650$\pm$0.1319}} & {17.23$\pm$2.67} & {0.6084$\pm$0.0903} &  {285.77} & {0.039} & {9.66} & {ViT} \\
        {\textbf{StegoGAN~\cite{Wu_2024_CVPR}}} & {\text{CVPR 2024}} & {-} & {0.4321$\pm$0.07953} & {16.76$\pm$2.02} & {\textbf{0.3084$\pm$0.0553}} & {106.32} & {11.37} & {227.46} & {GAN} \\
        {\textbf{CG-turbo~\cite{Parmar2024OneStepIT}}} & {\text{Arxiv 2024}} & {SD-turbo~\cite{Sauer2023AdversarialDD}} & {0.3207$\pm$0.0781} & {12.97$\pm$1.14} & {0.3429$\pm$0.0538} & {89.02} & {1540.28} & {16942.27} & {GAN} \\ \hline
        \textbf{LDM-f4($s=200, c_\theta=\mathcal{M}$)} & \text{CVPR 2022} & - & 0.3613$\pm$0.1470&  14.85$\pm$3.55&  0.4035$\pm$0.1146 & 108.98 & - & - & {DM} \\ 
        \textbf{LDM-f8($s=200, c_\theta=\mathcal{M}$)} & \text{CVPR 2022} & - & 0.4017$\pm$0.1391&  17.13$\pm$2.62& 0.3655$\pm$0.0935&  90.57 & {281.65} & {23813.62} & {DM} \\ 
        \textbf{PID-f8($s=200, c_\theta=\mathcal{M}$)} & \text{-} & - & 0.4006$\pm$0.1415 & 17.26$\pm$2.55 & 0.3599$\pm$0.0891&  \underline{84.68} & {281.65} & {23813.62} & {DM} \\ 
        {\textbf{PID-f8($s=5, c_\theta=\mathcal{E}$)}} & {-} & {-} & {\underline{0.4556$\pm$0.1350}} & {\textbf{18.16$\pm$2.52}} & {0.4245$\pm$01008} & {107.77} & {309.18} & {1939.11} & {DM} \\
        {\textbf{PID-f8($s=200,c_\theta=\mathcal{E}$)}} & {-} & {-} & {0.4115$\pm$0.0149} & {\underline{17.68$\pm$2.22}} & {\underline{0.3352$\pm$0.0867}} & {\textbf{78.00}} & {{309.18}} & {{24252.96}} & {{DM}} \\ \hline
    \end{tabular}
\end{table*}

\begin{figure*}[!ht]
    \centering 
    \includegraphics[width=1\linewidth]{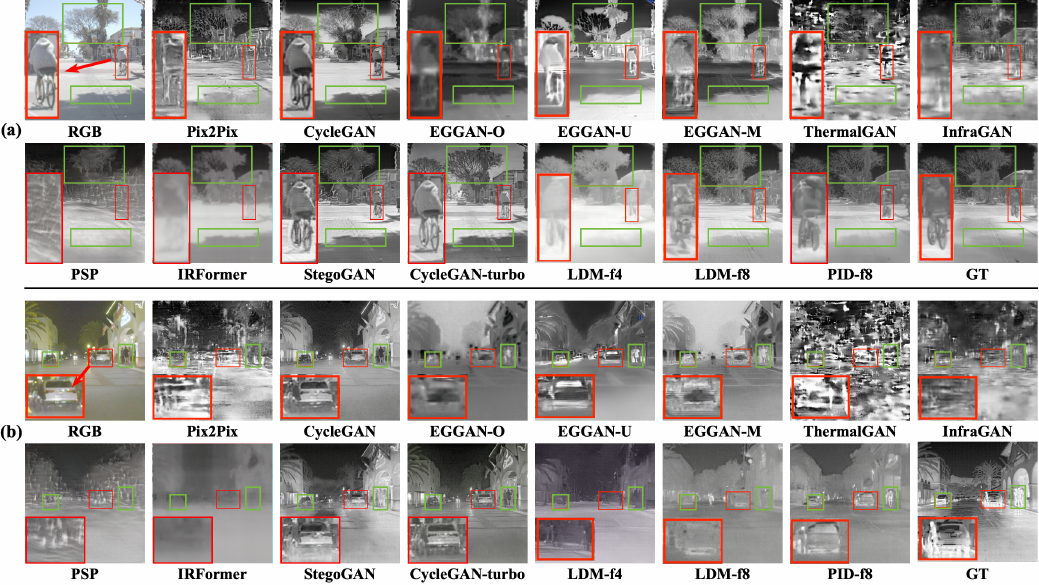}
    \caption{{Qualitative results on FLIR dataset. For the sake of clarity, the magnified region is highlighted with boxes for easier comparison. Our PID provides detailed depictions of components such as people and trees in the scene, accurately representing the important thermal objects.}}
    \label{fig:flir}
\end{figure*}

The FLIR dataset includes scenes of urban roads, highways, and other environments, featuring elements such as roads, buildings, pedestrians, and vehicles. Data was collected under both daytime and nighttime conditions. We compare the performance of various methods on the FLIR dataset, and the quantitative results are presented in \autoref{tab:flir}.

LDM-based methods achieve a balance between structural similarity and perceptual quality. Specifically, with fewer sampling steps (e.g., \(s=5\)), PID tends to exhibit smoothness, achieving high SSIM (\underline{0.4556}) and PSNR (\textbf{18.16}) scores, but it struggles to maintain perceptual quality, as indicated by inferior LPIPS and FID scores. When the number of sampling steps is increased (e.g., \(s=200\)), the perceptual quality converges to a superior level, with LPIPS of 0.3352 and FID of 78.00. This indicates a trade-off between generating globally smooth images and optimizing finer details. Overall, PID introduces consistent improvements in LPIPS and FID metrics, demonstrating its ability to generate images that align with the infrared domain while maintaining high perceptual quality.

However, the performance of LDM-based methods on the FLIR dataset is lower than that on the KAIST dataset. We attribute this disparity to pairing errors in the FLIR dataset. The public FLIR dataset contains mismatched image pairs, which negatively impact model training. In contrast, methods like StegoGAN~\cite{Wu_2024_CVPR} and CycleGAN-turbo~\cite{Parmar2024OneStepIT}, which are less sensitive to data pairing, exhibit relatively better performance.

As shown in \autoref{fig:flir}, while IRFormer achieves higher SSIM scores, it generates smoother images, resulting in lower perceptual quality. StegoGAN~\cite{Wu_2024_CVPR} preserves semantic features well, yielding strong LPIPS performance, but its similarity to the infrared domain is lower than PID, which achieves a 28.32 reduction in FID. These results highlight the diffusion model's ability to model rich distributions and the effectiveness of proposed physical losses in guiding image generation.

\autoref{fig:flir} also shows qualitative results on the FLIR dataset. CycleGAN retains RGB characteristics, but the translated images fail to capture thermal information (e.g., cyclist in \autoref{fig:flir}(a)). Edge-based methods like EGGAN~\cite{Lee2023EdgeguidedMR} produce clear edges but may not align with physical properties. For example, the tree shadow in \autoref{fig:flir}(a) has clear edges, but the infrared information of the shadow may contradict the actual physical situation. Our method, combining physical laws, generates more realistic infrared images.

In \autoref{fig:flir}(b), under low-light nighttime conditions, GAN-based methods produce blurry images or violate physical laws (e.g., cooler lights in EGGAN-O). LDM generates images similar to the ground truth but with some blurriness. Our proposed PID method generates clearer nighttime infrared images with strong robustness, as shown by the details of people and vehicles in \autoref{fig:flir}(b).

\subsubsection{Results on VEDAI dataset}

\label{sec:vedai}

\begin{table*}[!ht]
    \centering
    \caption{Quantitative metrics on VEDAI dataset. The best results are highlighted in \textbf{bold}, and the second-best results are \underline{underlined}. The rows above and below the line represent GAN-based methods and LDM-based methods, respectively. We set DDIM sampling step $s=20$.}
    \label{tab:vedai}
    \scriptsize
    \begin{tabular}{lccllllccc}
    \hline
        \textbf{Methods} & \makecell[c]{\textbf{Publication}\\\textbf{Venue}} & \makecell[c]{\textbf{Pre-}\\\textbf{trained}} & \textbf{SSIM}$\uparrow$ & \textbf{PSNR}$\uparrow$ & \textbf{LPIPS}$\downarrow$ & \textbf{FID}$\downarrow$ & \makecell[c]{\textbf{Param.}$\downarrow$\\\textbf{(M)}} & \makecell[c]{\textbf{MACs}$\downarrow$\\\textbf{(G)}} & \textbf{Type} \\ \hline
        \textbf{StegoGAN~\cite{Wu_2024_CVPR}} & \text{CVPR 2024} & - & \textbf{0.8091$\pm$0.0552} & 20.09$\pm$3.25&  0.1599$\pm$0.0543&  72.91 & 11.37 & 227.46 & GAN \\
        \textbf{CG-turbo~\cite{Parmar2024OneStepIT}} & \text{Arxiv 2024} & SD-turbo~\cite{Sauer2023AdversarialDD} & \underline{$0.6796\pm0.0882$} & 20.02$\pm$2.75&  0.1460$\pm$0.0507&  79.57 & 1540.28 & 16942.27 & GAN\\ \hline
        \textbf{LDM-f8 ($c_\theta=\mathcal{E}$)} & \text{CVPR 2022} & - & 0.6399$\pm$0.0910 &  \underline{25.27$\pm$2.37} & \underline{0.1061$\pm$0.0319} & \underline{70.91} & 309.18 & 3655.56 & DM \\ 
        \textbf{PID-f8 ($c_\theta=\mathcal{E}$)} & \text{-} & - & 0.6405$\pm$0.0922 & \textbf{25.39$\pm$2.27} & \textbf{0.1048$\pm$0.0304} & \textbf{69.80} & 309.18 & 3655.56 & DM \\ \hline
    \end{tabular}
\end{table*}

\begin{figure*}[tbp]
    \centering 
    \includegraphics[width=0.8\linewidth]{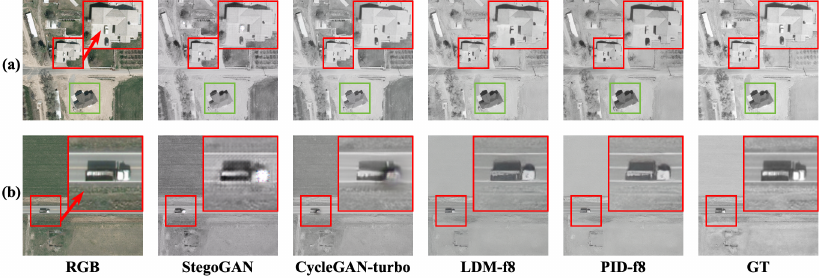}
    \caption{Qualitative results on VEDAI dataset. The visible RGB images and near-infrared images in the VEDAI dataset exhibit strong structural similarity. The PID method demonstrates good performance in generating details such as house features in (a) and vehicle details in (b).}
    \label{fig:vedai}
\end{figure*}

To further validate the effectiveness of the proposed PID method across different infrared wavelengths, we evaluate its performance on the VEDAI dataset, which contains paired aerial visible and near-infrared images. Given the high structural similarity between near-infrared and visible RGB images, we use the same latent space encoder $\mathcal{E}$ to map both images into a shared semantic space, preserving spatial similarity. The model converges after 30k training iterations on the VEDAI dataset.

The results, shown in \autoref{tab:vedai} and \autoref{fig:vedai}, demonstrate that PID outperforms state-of-the-art (SOTA) methods in PSNR, LPIPS, and FID metrics. While GAN-based methods like StegoGAN~\cite{Wu_2024_CVPR} achieve higher SSIM scores, they fall short of PID in perceptual quality (e.g., a 0.0551 reduction in LPIPS). GANs excel in smoothness due to the high structural similarity between visible and near-infrared images, enabling effective one-time transformations. In contrast, LDM and PID generate higher-frequency details with iterative sampling, sacrificing some smoothness. As shown in \autoref{fig:vedai}, PID provides clearer object details than baseline methods, which is critical for practical applications.

However, PID shows only slight improvement over LDM on the long-wavelength infrared dataset. This may be due to the TeV decomposition method, which is based on long-wavelength infrared theory and is less effective for near-infrared image generation.

\subsection{Ablation study}

\label{sec:ablation}

\subsubsection{$\mathcal{N}_{\text{TeV}}$ architecture}

\begin{table}[!htbp]
    \centering
    \caption{Training and testing $\mathcal{L}_{\text{Rec}}$ with different architectures of $\mathcal{N}_{\text{TeV}}$. Res50 means encoder of the net is resnet-50, res18 means resnet-18. $m$ represents the channels of \(\mathbf{V}\) in Eq.~\eqautoref{eq:VS}}.
    \scriptsize
    \begin{tabular}{ccccc}
    \hline
        Model & \textbf{Param.(M)} & \textbf{MACs(G)} & $\mathcal{L}_{\text{Train}}$ & $\mathcal{L}_{\text{Val}}$  \\ \hline
         PAN-res50, $m=4$ & 24.26 & 34.95 & 2.59e-5 & 3.03e-5  \\ 
         UNet-res18, $m=2$ & 14.33 & 21.82 & 8.70e-7 & 1.99e-6  \\
         UNet-res18, $m=4$ & 14.33 & 21.90 & 2.11e-6 & 2.12e-6  \\
         UNet-res18, $m=8$ & 14.33 & 22.05 & 1.77e-6 & 1.16e-6  \\ \hline
    \end{tabular}
    \label{tab:decom}
\end{table}

To determine a suitable framework for $\mathcal{N}_{\text{TeV}}$, we evaluate the reconstruction loss $\mathcal{L}_{\text{Rec}}$ across different frameworks. Both PAN (as used in \cite{Bao2023HeatassistedDA}) and UNet are tested using the KAIST dataset. All networks are trained for 950 epochs, with the framework and pretrained weights taken from \cite{Iakubovskii:2019}.

As shown in \autoref{tab:decom}, the UNet architecture consistently demonstrates lower reconstruction loss compared to PAN, indicating better convergence. Furthermore, UNet is more lightweight. Therefore, we select UNet with a ResNet-18 encoder as the architecture for $\mathcal{N}_{\text{TeV}}$.

\subsubsection{The effectiveness of $\mathcal{L}_{\text{Rec}}$ and $\mathcal{L}_{\text{TeV}}$} 

\begin{figure}[ht]
    \centering
    \includegraphics[width=1\linewidth]{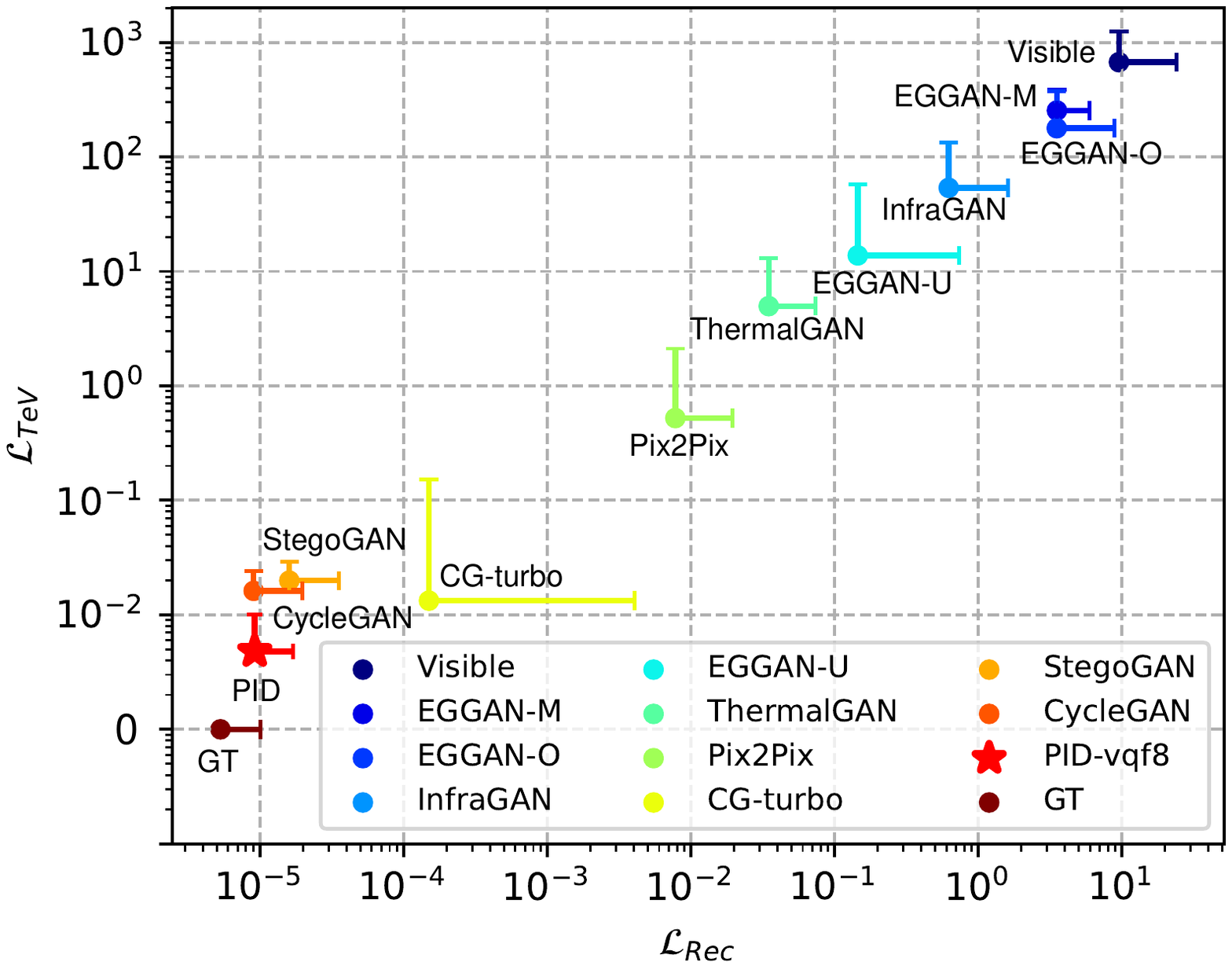}
    \caption{\(\mathcal{L}_{\text{Rec}}\) and \(\mathcal{L}_{\text{TeV}}\) of different generating images from KAIST test dataset. To make it clearer, we only plot the positive half of the error bars. Visible refers to visible RGB input, representing a high loss. Notably, PID shows remarkable results.}
    \label{fig:loss_quan}
\end{figure}

\begin{figure}[ht]
    \centering
    \includegraphics[width=1\linewidth]{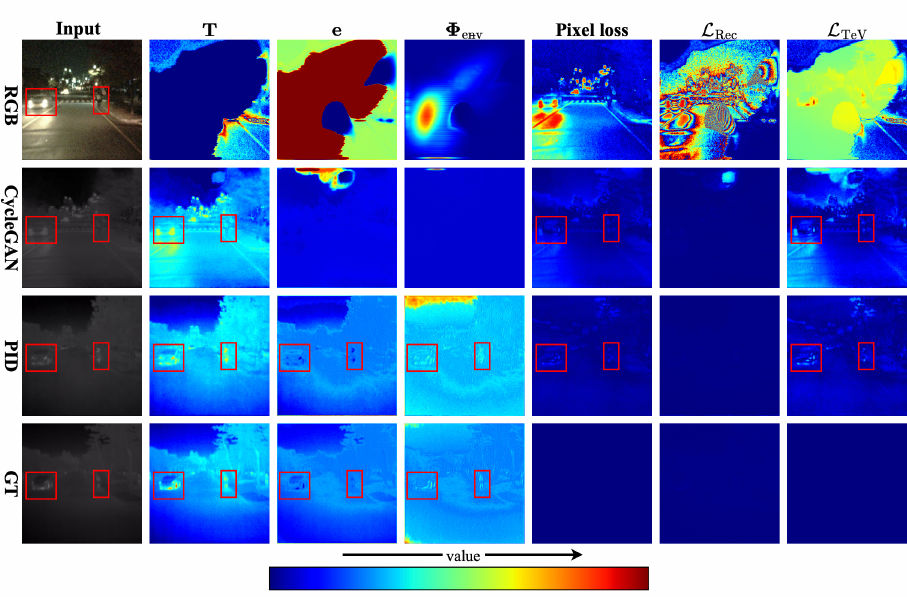}\\
    \caption{Visualization of $T$, $e$, and \(\Phi_{\text{env}}\), as well as pixel loss and proposed physical losses. {For clarity, we use pseudo-color to display both the TeV and the loss maps.} In PID and GT, the images show a clear distinction of pedestrians compared to the environment. Although the image translated by CycleGAN visually resembles the infrared image, it differs significantly from the GT in the TeV space. The proposed $\mathcal{L}_{\text{Rec}}$ and $\mathcal{L}_{\text{TeV}}$ provide a more accurate physical specificity than pixel loss.}
    \label{fig:lossmap}
\end{figure}

$\mathcal{L}_{\text{Rec}}$ quantifies the similarity between the translated infrared images and real infrared domain images. $\mathcal{L}_{\text{TeV}}$ measures the similarity between the translated images and the ground truth in the physical TeV space.

\autoref{fig:loss_quan} shows the performance of different generative methods and the ground truth images from the KAIST dataset, based on $\mathcal{L}_{\text{Rec}}$ and $\mathcal{L}_{\text{TeV}}$. RGB images exhibit the highest losses in both metrics, indicating poor alignment with the infrared domain. In contrast, PID achieves significantly lower losses than baseline methods, demonstrating its stronger capability to model the infrared domain.

\autoref{fig:lossmap} visualizes the TeV decomposition quantities. The main objects in the image—cars and a cyclist—are affected by glare and low-light conditions. CycleGAN fails to capture the thermal radiation of the car and cyclist, instead retaining the glare effects. In comparison, PID accurately models the temperature $\mathbf{T}$, emissivity $\mathbf{e}$, and $\Phi_{\text{env}}$ of objects, producing infrared images that are closer to the infrared domain and ground truth.

Additionally, \autoref{fig:lossmap} highlights the proposed losses. $\mathcal{L}_{\text{Rec}}$ acts as a discriminator for the infrared image domain, with higher losses observed for visible RGB images and lower losses for images closer to the real infrared ground truth. This highlights its ability to distinguish between visible and infrared image distributions. $\mathcal{L}_{\text{TeV}}$ highlights pixels that deviate from expected infrared physical properties, such as thermal radiation from tires and pedestrians. Overall, our proposed losses are explicitly targeted and highly effective, enhancing PID’s ability to generate physically accurate infrared images.

\subsubsection{Loss proportion}

\begin{table*}[htbp]
    \centering
    \caption{Ablation study on the loss proportion in PID. The best results are highlighted in \textbf{bold}, and the second-best results are \underline{underlined}.}
    \begin{tabular}{lllcccc}
    \hline
        \textbf{Dataset} & $\mathcal{L}_{\text{rec}}$ & $\mathcal{L}_{\text{TeV}}$ & \textbf{SSIM}$\uparrow$ & \textbf{PSNR}$\uparrow$ & \textbf{LPIPS}$\downarrow$ & \textbf{FID}$\downarrow$  \\ \hline
        & \XSolidBrush & \XSolidBrush & 0.7892 & 23.31 & 0.1380 & 64.54  \\
        & \XSolidBrush & \CheckmarkBold($k_2=50$) & 0.7892 & 23.53 & \underline{0.1366} & 50.99 \\ 
        \textbf{KAIST} & \CheckmarkBold($k_1=50$) & \XSolidBrush & \underline{0.7901} &  \textbf{23.63} & \textbf{0.1363} & \textbf{50.17}  \\ 
        & \CheckmarkBold($k_1=50$) & \CheckmarkBold($k_2=50$) & 0.7893 & 23.43 & 0.1370 & \underline{50.66}  \\
        & \CheckmarkBold($k_1=50$) & \CheckmarkBold($k_2=5$) & \textbf{0.7913} & \underline{23.60} & \underline{0.1366} & 51.69  \\ \hline
        \multirow{2}{*}{\textbf{FLIR}} & \CheckmarkBold($k_1=50$) & \CheckmarkBold($k_2=50$) & \textbf{0.4006} & \underline{17.26} & \underline{0.3599} & \textbf{84.26} \\ 
        ~ & \CheckmarkBold($k_1=50$) & \CheckmarkBold($k_2=5$) & \underline{0.3991} & \textbf{17.35} & \textbf{0.3577} & \underline{84.28} \\ 
        \hline
    \end{tabular}
    \label{tab:abalation-loss}
\end{table*}

This section presents an ablation study on the proportion of $\mathcal{L}_{\text{Rec}}$ and $\mathcal{L}_{\text{TeV}}$. We test five different proportions: $\left\{ k_{1}, k_{2} | k_{1} \in \{0, 50\}, k_{2} \in \{0, 50\} \right\}$ and $\{ k_1=50, k_2=5 \}$. \autoref{tab:abalation-loss} summarizes the evaluation of translated image quality. PID with only $\mathcal{L}_{\text{Rec}}$ or $\mathcal{L}_{\text{TeV}}$ (second and third rows) outperforms the baseline LDM, confirming the effectiveness of our proposed losses. As shown in \autoref{tab:abalation-loss}, $k_1=50, k_2=5$ (fifth row) performs better than $k_1=50, k_2=50$ (fourth row). A slightly lower proportion of $\mathcal{L}_{\text{TeV}}$ improves model performance, likely due to reduced overfitting. Since translating infrared images from RGB images is an underdetermined problem, an excessive dependence on $\mathcal{L}_{\text{TeV}}$ may lead to overfitting, where the model overemphasizes the training data distribution and overlooks other plausible distributions.

\subsubsection{Dimensions of \( \mathbf{V} \)}

\begin{table}[htbp]
    \centering
    \caption{Ablation study on the hyperparameter \(m\) in Eq.(\ref{eq:VS}). The best results are highlighted in \textbf{bold}, and the second-best results are \underline{underlined}.}
    \begin{tabular}{ccccc}
    \hline
        $\mathbf{V}$ & \textbf{SSIM}$\uparrow$ & \textbf{PSNR}$\uparrow$ & \textbf{LPIPS}$\downarrow$ & \textbf{FID}$\downarrow$  \\ \hline
         $m=2$ & 0.7861 & 23.33 & \underline{0.1371} & \underline{51.26}  \\ 
         $m=4$ & \textbf{0.7893} & \underline{23.43} & \textbf{0.1370} & \textbf{50.66}  \\
         $m=8$ & \underline{0.7885} & \textbf{23.51} & 0.1392 & 51.73  \\ \hline
    \end{tabular}
    \label{tab:ablation-V}
\end{table}

In this section, we conduct an ablation study on the hyperparameter $ m $, which represents the number of channels used to estimate $ \Phi_{\text{env}} $ in Eq.~\eqautoref{eq:VS}. We train $ \mathcal{N}_{\text{TeV}} $ with $ m \in \{2, 4, 8\} $, embedding each into the PID framework. All experiments are conducted using the KAIST dataset with the loss proportions set to $ k_1 = k_2 = 50 $.

\autoref{tab:ablation-V} presents the results of translated infrared images. We find that both excessively high and low values of $ m $ are unsuitable for PID training. When $ m = 2 $, the limited number of decomposition channels in $ \mathbf{V} $ leads to poor estimations of $ \Phi_{\text{env}} $, resulting in inferior generative performance. On the other hand, with $ m = 8 $, the increased channels may cause overfitting, negatively impacting the overall PID training. When $ m = 4 $, the number of downsampling grids aligns well with the four quadrants of the image, yielding the best performance. Thus, we select $ m = 4 $ for PID.

\subsubsection{Comparison with other physical decomposition}

We compare the TeV method with the TeS~\cite{Gillespie1998ATA} method, which is designed for remote sensing infrared images. Both methods are implemented using the same neural network, but differ in their outputs. The TeS network directly estimates $ \Phi_{\text{env}} $, with the output $ \mathcal{N}_{\text{TeS}}(\mathcal{S}) = \mathbf{e}, \mathbf{T}, \Phi_{\text{env}} $.

As shown in \autoref{tab:tes}, on the KAIST dataset, TeS exhibits slower convergence compared to TeV. This is because TeS relies on a single-step estimation of $ \Phi_{\text{env}} $, which complicates the model. In contrast, TeV uses multi-directional joint estimation, better reflecting real-world heat transfer laws, simplifying the process and leading to faster convergence. 

Additionally, we compute the Earth Mover's Distance~\cite{rubner2000earth} (EMD) between the reconstruction loss distributions, \( p(\mathcal{L}_{\text{Rec}}(\boldsymbol{x}_{\text{RGB}})) \) for the visible domain and \( q(\mathcal{L}_{\text{Rec}}(\boldsymbol{x}_{\text{IR}})) \) for the infrared domain. EMD quantifies the difference between two distributions by minimizing the cost of transforming one to match the other. A larger \( \text{EMD}(p,q) \) indicates a greater distribution difference, reflecting improved domain discrimination. The TeV method leverages the relationship between neighboring pixels in the image (exhibited in $\Phi_{\text{env}}$), which is crucial for infrared imaging due to its distinct spatial patterns and correlations compared to visible RGB images. In contrast, the TeS method neglects this relationship, leading to performance degradation in RGB images as it fails to distinguish effectively between the visible and infrared domains. The \( \mathcal{L}_{\text{Rec}} \) loss function in TeVNet improves domain discrimination, enhancing PID training.

\begin{table}[!ht]
    \centering
    \caption{Comparison of $\mathcal{L}_{\text{Rec}}$ on the KAIST dataset. $ p(\mathcal{L}_{\text{Rec}}(\boldsymbol{x}_{\text{RGB}})) $ and $ q(\mathcal{L}_{\text{Rec}}(\boldsymbol{x}_{\text{IR}})) $ represent the loss distributions of visible and infrared images, respectively. TeVNet shows better convergence and superior domain discrimination between visible and infrared images.}
    \label{tab:tes}
    \begin{tabular}{cccc}
    \hline
        \textbf{Methods} & \textbf{IR Train} & \textbf{IR Test} & $\text{EMD}(p,q)$ \\ \hline
        \textbf{TeSNet} & $4.00 \times 10^{-6}$ & $3.99 \times 10^{-6}$ & 0.17 \\ 
        \textbf{TeVNet} & $2.11 \times 10^{-6}$ & $2.12 \times 10^{-6}$ & 9.54 \\ \hline
    \end{tabular}
\end{table}

\subsubsection{Complexity and sampling steps analysis}
 
The inference pf LDM/PID consists of three key components: the conditioner $ c_\theta $, the UNet sampler, and the decoder $ \mathcal{D} $. The total MACs (Multiply-Accumulate operations) for the LDM/PID are given by:
\begin{equation}
    \text{MACs}(\text{LDM/PID}) = \text{MACs}(c_\theta) + \text{MACs}(\text{UNet}) \times s + \text{MACs}(\mathcal{D}),
\end{equation}
where $ \text{MACs}(c_\theta = \mathcal{E}) = 439.34 \, \text{G} $, $ \text{MACs}(\text{UNet}) = 114.43 \, \text{G} $, and $ \text{MACs}(\mathcal{D}) = 927.62 \, \text{G} $. The sampling cost increases linearly with the number of steps $ s $. As shown in Tables \ref{tab:kaist}, \ref{tab:flir}, and \ref{tab:vedai}, for most datasets, fewer than 20 steps suffice for good performance. While the sampling cost for 20 steps is higher than that of lightweight GAN networks, it is still more computationally efficient than large GAN models.

We compare the computational complexity of the PID method with CycleGAN-turbo across various datasets and assess the effect of different sampling steps using DDIM~\cite{Song2020DenoisingDI}. As demonstrated in \autoref{tab:ddim_steps}, PID with DDIM achieves a balance between image quality and computational cost. For the KAIST dataset, with just 4 steps ($ s = 4 $), PID achieves an LPIPS of 0.1681, which is comparable to CycleGAN-turbo's performance, while consuming only 1385.34 GMACs, about one-tenth of CycleGAN-turbo's cost. The PID method with DDIM sampling is flexible, allowing adjustments between precision and speed to meet the specific needs of various datasets.

\begin{table*}[!ht]
    \centering
    \caption{Ablation study on different sampling steps $ s $ with DDIM. We compare LPIPS and computational costs (MACs) of the PID method and CycleGAN-turbo on the KAIST and VEDAI datasets. The PID method offers a better balance between image generation quality and sampling speed.}
    \scriptsize
    \begin{tabular}{ccccccccccc}
    \hline
        ~ & \multirow{2}{*}{\textbf{Dataset}} & \multirow{2}{*}{\textbf{$c_\theta$}} & ~ & \multicolumn{5}{c}{\textbf{Sampling steps}} & ~ & \multirow{2}{*}{\textbf{CycleGAN-turbo}}\\ 
        \cline{4-10}
        ~ & ~ & ~ & \textbf{$s=2$} & \textbf{$s=4$} & \textbf{$s=5$} & \textbf{$s=10$} & \textbf{$s=20$} & \textbf{$s=50$} & \textbf{$s=100$}  & ~ \\ \hline
        \textbf{LPIPS}$\downarrow$ & \multirow{2}{*}{\textbf{KAIST}} & \multirow{2}{*}{$\mathcal{M}$} & 0.2062 & 0.1681 & 0.1615 & 0.1463 & 0.1389 & \textbf{0.1365} & 0.1371 & 0.1947 \\  
        \textbf{MACs(G)}$\downarrow$ & ~ & ~ & 1156.48 & 1385.34 & 1499.77 & 2071.92 & 3216.22 & 6649.12 & 12370.62 & 16942.27 \\ \hline
        \textbf{LPIPS}$\downarrow$ & \multirow{2}{*}{\textbf{VEDAI}} & \multirow{2}{*}{$\mathcal{E}$} & 0.2608 & 0.1610 & 0.1441 & 0.1086 & \textbf{0.1048} & 0.1101 & 0.1132 & 0.1460 \\ 
        \textbf{MACs(G)}$\downarrow$ & ~ & ~ & 1595.82 & 1824.68 & 1939.11 & 2511.26 & 3655.56 & 7088.46 & 12809.96 & 16942.27 \\ \hline
    \end{tabular}
    \label{tab:ddim_steps}
\end{table*}

\subsubsection{Images prior denoising}

\begin{table}[!ht]
    \centering
    \caption{{Ablation study on prior image denoising. For the visible RGB images guidance, we apply NLM denoising before inputting them into the diffusion model for guided sampling. DN=Denoising. $c_\theta=\mathcal{M},s=200$.}}
    \scriptsize
    \begin{tabular}{ccccc}
    \hline
        \textbf{} & \multicolumn{2}{c}{{\textbf{FLIR Day}}} & \multicolumn{2}{c}{{\textbf{FLIR Night}}}  \\ 
        \cline{2-3} \cline{4-5}
        \textbf{} & {\textbf{PSNR}$\uparrow$} & {\textbf{LPIPS}$\downarrow$} & {\textbf{PSNR}$\uparrow$} & {\textbf{LPIPS}$\downarrow$} \\ \hline
        {\textbf{PID w/o DN}}  & \textbf{18.72$\pm$1.94} & \textbf{0.3305$\pm$0.0927} & 14.98$\pm$1.48 & \textbf{0.4060$\pm$0.0585}  \\ 
        {\textbf{PID w DN}} & 18.56$\pm$2.14 & 0.3496$\pm$0.0988 & \textbf{15.19$\pm$1.34} & 0.4198$\pm$0.0575  \\ \hline
    \end{tabular}
    \label{tab:denoising}
\end{table}

The FLIR night dataset contains noticeable noise, prompting us to conduct prior image denoising experiments. Specifically, we apply Non-Local Means denoising~\cite{1467423} to both day and night visible RGB images before inputting them into the diffusion model. The results, presented in \autoref{tab:denoising}, highlight the impact of denoising on image quality. For the night subset, denoising improves the PSNR of translated images by reducing noise and enhancing smoothness. However, the LPIPS worsens, indicating a trade-off in perceptual quality. In contrast, for the day subset, denoising provides no such benefit. Denoising is effective when the input image contains substantial noise (e.g., \autoref{fig:flir}(b)). When the image is relatively clean, applying denoising can blur fine details, thereby degrading the quality of the generated image.

\section{Discussions}
\label{sec:discussion}

This section provides more analysis of the proposed method, addressing computational cost, sampling stability, and challenges in real-world applications.

\subsection{Computational cost}
\label{sec:discussion-1}

The computational cost of the proposed method increases with the number of sampling steps during inference. To mitigate this, we use DDIM~\cite{Song2020DenoisingDI} to accelerate inference, achieving faster sampling with only 22\% of the computational cost of CycleGAN-Turbo, while maintaining high-quality image generation (LPIPS$\downarrow$ -34\%) on the KAIST dataset. However, there is still room for further optimization, especially when compared to lightweight GAN-based methods. Future work will focus on accelerating the sampling process to strike an optimal balance between image quality and computational efficiency.

\subsection{Sampling stability}
\label{sec:discussion-2}

The sampling process may introduce redundant details due to fluctuations during iterative sampling, particularly in visible-to-infrared image translation, which involves large modality differences. To address this, we improve sampling stability by selecting an appropriate number of steps and setting the DDIM hyperparameter $\eta=0$. Future research will focus on better constraining image details and improving the semantic coherence of the generated images.

\subsection{Generalization to real-world scenarios}
\label{sec:discussion-3}

Our TeV decomposition method, based on long-wavelength infrared (LWIR) theory, demonstrates notable improvements on LWIR datasets such as KAIST and FLIR. However, its performance on near-infrared datasets like VEDAI is less pronounced, suggesting the need for further investigation into the physical constraints of near-infrared image generation. 

Additionally, RGB images captured under low-light conditions challenge the model's ability to extract semantic information, leading to lower-quality generated infrared images. As observed in \autoref{tab:denoising}, prior image denoising improves night-time image quality, but results remain suboptimal. Future research will focus on enhancing the model's ability to extract meaningful features from such low-quality input images.

\section{Conclusion}
\label{sec:conclusion}

This paper presents a novel visible-to-infrared image translation method, with the primary contribution being the integration of physical laws into the training of a latent diffusion model. We introduce the TeV decomposition method for spectral integration, which provides physical losses during the diffusion model training process, ensuring that the generated results align with physical laws. Extensive experiments demonstrate that our method outperforms existing approaches. We believe this technique has significant potential for infrared data augmentation and could inspire further research into how generative models capture real-world physical phenomena. In the future work, we will focus on enhancing the generative capacity of the model while also reducing its computational cost.

\appendix
\setcounter{figure}{0}
\setcounter{table}{0}
\section{Diffusion model derivation}
\label{app1}
	The posterior distribution of the image after adding noise at different process steps can be written as the product of the posterior distributions:
	\begin{equation}
		q\left( \boldsymbol{x}_{1:T}|\boldsymbol{x}_{0}\right)  =\prod\nolimits^{T}_{t=1} q\left( \boldsymbol{x}_{t}|\boldsymbol{x}_{t-1}\right).
	\end{equation}
	
	In the diffusion model, the noise diffusion process is considered as a Markov chain~\cite{Ho2020DenoisingDP}. The conditional probability distribution of the image after noise diffusion and the image in the previous process step is assumed to follow a Gaussian distribution, independent of earlier image distributions. The variance of noise in each iteration is represented by the scalar hyper-parameter \(\alpha_{1:T}\), subjected to \(0<\alpha_t<1\).
	\begin{equation}
		\begin{aligned}
			q\left( \boldsymbol{x}_{t}|\boldsymbol{x}_{t-1}\right)  &=\mathcal{N}\left( \boldsymbol{x}_{t}|\sqrt{\alpha_{t}}\boldsymbol{x}_{t-1},\left( 1-\alpha_{t} \right)  \mathbf{I} \right).\\
			\boldsymbol{x}_{t}&=\sqrt{\alpha_{t}} \boldsymbol{x}_{t-1}+\sqrt{1-\alpha_{t}} \boldsymbol{\epsilon},\\
			\boldsymbol{\epsilon} &\sim \mathcal{N}\left( \mathbf{0} ,\mathbf{I} \right).
		\end{aligned}
	\end{equation}
	
	The conditional probability distribution between different process step can be uniformly represented using the initial noise-free image \(\boldsymbol{x}_{0}\) and the parameter \(\bar{\alpha}_t\) corresponding to each stage.
	\begin{equation}
		\begin{aligned}
			\boldsymbol{x}_{t}&=\sqrt{\alpha_{t}} \boldsymbol{x}_{t-1}+\sqrt{1-\alpha_{t}} \boldsymbol{\epsilon}\\
			&=\sqrt{\alpha_{t}} \left( \sqrt{\alpha_{t-1}} \boldsymbol{x}_{t-2}+\sqrt{1-\alpha_{t-1}} \boldsymbol{\epsilon} \right)  +\sqrt{1-\alpha_{t}} \boldsymbol{\epsilon} \\
			&=\sqrt{\alpha_{t} \alpha_{t-1}} \boldsymbol{x}_{t-2}+\sqrt{1-\alpha_{t} \alpha_{t-1}} \boldsymbol{\epsilon}\\
			&=\sqrt{\alpha_{t} \alpha_{t-1} ...\alpha_{1}} \boldsymbol{x}_{0}+\sqrt{1-\alpha_{t} \alpha_{t-1} ...\alpha_{1}} \boldsymbol{\epsilon}.
		\end{aligned}
	\end{equation}
	This conditional probability distribution is also Gaussian distribution:
	\begin{equation}
		\begin{aligned}
			q\left( \boldsymbol{x}_{t}|\boldsymbol{x}_{0}\right) &=\mathcal{N}\left( \boldsymbol{x}_{t}|\sqrt{\bar{\alpha}_{t}} \boldsymbol{x}_{0},\left( 1-\bar{\alpha}_{t} \right)  \mathbf{I} \right),\\
			\bar{\alpha}_t  &= \prod\nolimits^{t}_{i=1} \alpha_i.
		\end{aligned}
	\end{equation}
	
	Given the image distribution of \(\boldsymbol{x}_{0}\) and \(\boldsymbol{x}_{t}\), the posterior distribution of \(\boldsymbol{x}_{t-1}\) can be derived with some algebraic manipulation.
	\begin{subequations}
		\begin{align}
			q(&{\boldsymbol{x}}_{t-1}|\boldsymbol{x}_0,\boldsymbol{x}_t) = \mathcal N(\boldsymbol{x}_{t-1}|\boldsymbol{\mu} ,\sigma^2 \mathbf{I}).\\
			\label{q(yt-1,y0,yt)b}\boldsymbol{\mu} &=\frac{\sqrt{\bar{\alpha}_{t-1}} \left( 1-\alpha_{t} \right)}{1-\bar{\alpha}_{t}} \boldsymbol{x}_{0}+\frac{\sqrt{\alpha_{t}} \left( 1-\bar{\alpha}_{t-1} \right)}{1-\bar{\alpha}_{t}} \boldsymbol{x}_{t}.\\
			\sigma^{2} &=\frac{(1-\bar{\alpha}_{t-1} )(1-\alpha_{t} )}{1-\bar{\alpha}_{t}}.
		\end{align}
		\label{q(yt-1,y0,yt)}
	\end{subequations}
	The q-process, which represents the noise diffusion process, is easily understood in the above equation. However, in the actual inference process (from \(\boldsymbol{x}_{T}\) to \(\boldsymbol{x}_{0}\)), \eqautoref{q(yt-1,y0,yt)b} requires prior knowledge of the image distribution \(\boldsymbol{x}_{0}\) at \(t=0\) (i.e., the desired high-quality image), which contradicts the purpose of image generation. Therefore, additional work is required to optimize the denoising process
	
	Given \(t\), the noise-diffused image and the image at \(t=0\) can be described based on \eqautoref{y,gamma}.
	\begin{equation}
		\boldsymbol{x}_t =\sqrt{\bar{\alpha}_t} \boldsymbol{x}_{0}+\sqrt{1-\bar{\alpha}_t} \boldsymbol{\epsilon} , \boldsymbol{\epsilon} \sim \mathcal{N}\left( \mathbf{0} ,\mathbf{I} \right).
		\label{y,gamma}
	\end{equation}
	Therefore, a denoising model \(\boldsymbol{\epsilon}_\theta\) can be trained to fit the noise at different diffusion process step. 
	\begin{equation}
		\mathbb{E}_{\left( \boldsymbol{x} \sim p_{data}\right)} \mathbb{E}_{\boldsymbol{\epsilon} \sim \mathcal{N}(\mathbf{0},\mathbf{I})} \parallel \boldsymbol{\epsilon}_{\theta}\left( \boldsymbol{x}_t; t,\mathbf{c} \right)  -\boldsymbol{\epsilon} \parallel^{p}_{p}.
	\end{equation}
	The denoising process can be written as \eqautoref{p-process}.
	\begin{subequations}
		\begin{align}
			p_{\theta}\left( \boldsymbol{x}_{0:T}|\boldsymbol{x}\right)  =p\left( \boldsymbol{x}_{T}\right) & \prod\nolimits^{T}_{t=1} p_{\theta}\left( \boldsymbol{x}_{t-1}|\boldsymbol{x}_{t},\mathbf{c}\right),\\
			p\left( \boldsymbol{x}_{T}\right)  &=\mathcal{N}\left( \boldsymbol{x}_{T}|\mathbf{0} ,\mathbf{I} \right),\\
			p_{\theta}\left( \boldsymbol{x}_{t-1}|\boldsymbol{x}_{t},\mathbf{c}\right)  =\mathcal{N}&\left( \boldsymbol{\mu}_{\theta} \left( \boldsymbol{x}_{t};t,\mathbf{c} \right)  ,\sigma^{2}_{t} \mathbf{I} \right).
		\end{align}
		\label{p-process}
	\end{subequations}
	By revisiting the definition of \(\boldsymbol{x}_t\), the denoising model can estimate the noise and obtain the distribution of the image at \(t=0\), denoted as \(\hat{\boldsymbol{x}_{0}}\), thus avoiding the need for prior knowledge of the image distribution at \(t=0\).
	\begin{equation}
		\hat{\boldsymbol{x}_{0}} =\frac{1}{\sqrt{\bar{\alpha}_{t}}} \left( \boldsymbol{x}_{t}-\sqrt{1-\bar{\alpha}_{t}} \boldsymbol{\epsilon}_{\theta}\left( \boldsymbol{x}_t; t,\mathbf{c} \right)  \right).
	\end{equation}
	By substituting the estimated \(\hat{\boldsymbol{x}_0}\) into \eqautoref{q(yt-1,y0,yt)b}, the mean \(\mu_{\theta}\) representing the p-process can be obtained. In this case, the variance \(\sigma^2\) of the normal distribution remains unchanged.
	\begin{equation}
		\boldsymbol{\mu}_{\theta} \left( \boldsymbol{x}_{t};t, \mathbf{c} \right)  =\frac{1}{\sqrt{\alpha_{t}}} \left( \boldsymbol{x}_{t}-\frac{1-\alpha_{t}}{\sqrt{1-\bar{\alpha}_{t}}} \boldsymbol{\epsilon}_{\theta}\left( \boldsymbol{x}_t; t,\mathbf{c} \right)  \right).
		\label{mu-theta}
	\end{equation}
	Based on \eqautoref{q(yt-1,y0,yt)} and \eqautoref{mu-theta}, the iteration of inverse denoising process can be written as \eqautoref{equation-iteration}.
	\begin{equation}
		\begin{aligned}
			\boldsymbol{x}_{t-1}\leftarrow \frac{1}{\sqrt{\alpha_{t}}} ( \boldsymbol{x}_{t}-\frac{1-\alpha_{t}}{\sqrt{1-\bar{\alpha}_{t}}} \boldsymbol{\epsilon}_{\theta}\left( \boldsymbol{x}_t; t,\mathbf{c} \right)) \\
			+\sqrt{\frac{1-\bar{\alpha}_{t-1}}{1-\bar{\alpha}_{t}} (1-\alpha_{t} )} \boldsymbol{\epsilon},  \boldsymbol{\epsilon} \sim \mathcal{N}\left( \mathbf{0} ,\mathbf{I} \right).
		\end{aligned}
		\label{equation-iteration}
	\end{equation}
	The iterative denoising process starts with \(\boldsymbol{x}_{T} \sim \mathcal{N}\left( \mathbf{0} ,\mathbf{I} \right)\) and repeats the iterations according to \eqautoref{equation-iteration} until \(t=0\) to obtain a high-quality generation result \(\boldsymbol{x}_0\).

\section{Emissivity of common materials}
\label{app2}

\begin{figure}[!htbp]
    \centering
    \includegraphics[width=0.6\linewidth]{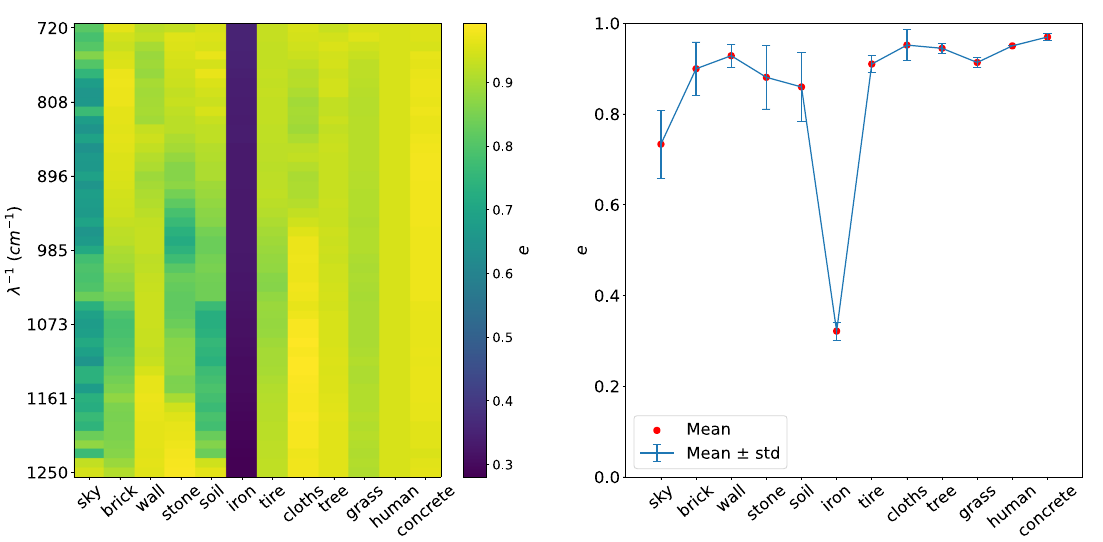}
    \caption{Common materials and their emissivities at different wavelengths. $e$ represents emissivity, $\lambda$ represents wavelength. HADAR~\cite{Bao2023HeatassistedDA} collects the emissivities of some of the common materials in the figure. Left: the emissivities of different materials at different wavelengths. Right: the statistical result of the emissivities of different materials. Both figures indicate that emissivity of a special materials changes little at different wavelengths.}
    \label{fig:matlib}
\end{figure}

\autoref{fig:matlib} shows the emissivities of common materials as it varies with wavelength, with data sourced from HADAR~\cite{Bao2023HeatassistedDA}. According to the statistical results, we observe that the emissivities of most materials changes little with wavelength. Therefore, it can be approximately considered that specific materials have a constant emissivity.

\section{Implementation details}
\label{app3}

\subsection{Integration of TeV decomposition}

\autoref{al:tevnet} provides pseudo-code for integrating TeV decomposition during the diffusion model training process.

\begin{algorithm}[!h]
    \caption{Integration of TeV decomposition}
    \label{al:tevnet}
    \begin{algorithmic}[1]
    \item \textbf{Input}: $\boldsymbol{x}_0, \boldsymbol{x}_{\text{RGB}} \sim \text{Dataset}$, $\boldsymbol{\epsilon} \sim \mathcal{N}(\mathbf{0},\mathbf{I})$.
    
    \item $\boldsymbol{z}_0 = \mathcal{E}(\boldsymbol{x}_{0})$, $\mathbf{c}_{\text{RGB}} = c_\theta (\boldsymbol{x}_{\text{RGB}})$.
    
    \item $\boldsymbol{z}_t = \sqrt{\bar{\alpha}_t}\boldsymbol{z}_0+\sqrt{1-\bar{\alpha}_t}\boldsymbol{\epsilon}$.
    
    \item $\hat{\boldsymbol{x}_{0}} = \mathcal{D} \left( \frac{1}{\sqrt{\bar{\alpha_{t}}}} ({\boldsymbol{z}_{t} - \sqrt{1-\bar{\alpha_{t}}} \boldsymbol{\epsilon}_{\theta}(\boldsymbol{z}_{t};t,\mathbf{c}_{\text{RGB}})})\right)$.
    
    \item $\mathcal{L}_{\text{Noise}}=\mathcal{L}_1(\boldsymbol{\epsilon},\boldsymbol{\epsilon}_{\theta})$.
    
    \item $\mathcal{L}_{\text{Rec}} = \mathcal{L}_\text{MSE}(\text{Rec} (\mathcal{N}_{\text{TeV}} (\hat{\boldsymbol{x}}_{0} )), \hat{\boldsymbol{x}}_{0})$.
    
    \item $\mathcal{L}_{\text{TeV}} = \mathcal{L}_\text{MSE}(\mathcal{N}_{\text{TeV}} (\hat{\boldsymbol{x}}_{0} ),\mathcal{N}_{\text{TeV}} (\boldsymbol{x}_{0}))$.
    
    \item Calculate loss $\mathcal{L} = \mathcal{L}_{\text{Noise}}+k_1\mathcal{L}_{\text{Rec}}+k_2\mathcal{L}_{\text{TeV}}$.
    
    \item Calculate gradient $\nabla_{\theta} \mathcal{L} $ and gradient back propagation.
    
    \item until converged.
    \end{algorithmic}
\end{algorithm}

\subsection{Hyperparameters}

\autoref{tab:achi} provides the network architecture, hyperparameters, and training schedule to ensure reproducibility.

\begin{table}[!ht]
    \centering
    \caption{Hyperparameters for the models. DMs are trained on 4 NVIDIA V100. TeVNets are trained on a single NVIDIA V100.}
    \label{tab:achi}
    \scriptsize
    \begin{tabular}{llccc}
    \hline
        \textbf{Model} & \textbf{Dataset} & \textbf{KAIST} & \textbf{FLIR} & \textbf{VEDAI} \\ \hline
        \multirow{16}{*}{LDM} & $\boldsymbol{z}$ shape & $64\times64\times4$ & $64\times64\times4$ & $64\times64\times4$ \\ 
        ~ & Diffusion steps & 1000 & 1000 & 1000 \\ 
        ~ & Noise schedule & linear & linear & linear \\ 
        ~ & Channels & 128 & 128 & 128 \\ 
        ~ & Depth & 2 & 2 & 2 \\ 
        ~ & Attention resolution & 32,16,8 & 32,16,8 & 32,16,8 \\ 
        ~ & Channel multiplier & 1,2,4 & 1,2,4 & 1,2,4 \\ 
        ~ & Numbers of heads & 8 & 8 & 8 \\ 
        ~ & Batch size & 48 & 48 & 48 \\
        ~ & Accumulate batch & 2 & 2 & 2 \\
        ~ & Iterations & 60k & 60k & 30k \\ 
        ~ & Learning rate & 1e-6 & 1e-6 & 1e-6 \\ 
        ~ & Optimizer & AdamW & AdamW & AdamW \\ 
        ~ & Pretrained & None & None & None \\ 
        ~ & Default sampling steps & 20 & 200 & 20 \\ 
        ~ & DDIM $\eta$ & 0 & 0 & 0 \\ \hline
        \multirow{5}{*}{TeVNet} & Batch Size & 64 & 64 & 64 \\ 
        ~ & Learning rate & 1e-3 & 1e-3 & 1e-3 \\
        ~ & Epochs  & 950 & 1k & 450 \\ 
        ~ & Optimizer & Adam & Adam & Adam \\
        ~ & Pretrained & ImageNet & ImageNet & ImageNet \\ \hline
    \end{tabular}
\end{table}

\section*{Acknowledgements}

This work was supported in part by the National Natural Science Foundation of China under Grant No. U23B2034, No. 62203424, and No. 62176250, and in part by the Innovation Program of Institute of Computing Technology, Chinese Academy of Sciences under Grant No. 2024000112.

\scriptsize
\bibliographystyle{elsarticle-num} 
\bibliography{main.bib}
\end{document}